\documentclass[12 pt, letterpaper]{article}
\usepackage[verbose=true,letterpaper]{geometry}
\usepackage[utf8]{inputenc} % allow utf-8 input
\usepackage[T1]{fontenc}    % use 8-bit T1 fonts
\usepackage{hyperref}       % hyperlinks
\usepackage{url}            % simple URL typesetting
\usepackage{booktabs}       % professional-quality tables
\usepackage{amsfonts,amsmath,amssymb,amsthm} % various math stuff}       % blackboard math symbols
\usepackage{nicefrac}       % compact symbols for 1/2, etc.
\usepackage{microtype}      % microtypography
\usepackage{xcolor}         % colors
\usepackage{microtype}
\usepackage{graphics}       % figures
\usepackage{graphicx}
\usepackage{subfigure}
\usepackage{booktabs}       % for professional tables
\usepackage{textcomp}
\usepackage{sidecap}  % For putting caption beside figure 
\usepackage{algorithm}
\usepackage[noend]{algpseudocode}
\algrenewcommand{\algorithmiccomment}[1]{\hskip1em$\rightarrow$ #1}
\newcommand{\sample}{\text{sample }}
\usepackage{babel}
\usepackage{apalike}
\newcommand{\E}{\mathbb{E}}
\newcommand{\Dat}{\mathcal{D}}
\newcommand{\inv}{^{-1}}
\newcommand{\trp}{{^\top}} % transpose

\newcommand{\Tr}{\mathrm{Tr}}
\renewcommand{\inv}{^{-1}}
\newcommand{\diag}{\mathrm{diag}}
\newcommand{\vw}{\mathbf{w}}
\newcommand{\vx}{\mathbf{x}}

\newcommand{\vzero}{\mathbf{0}}
\newcommand{\Nrm}{\mathcal{N}}  % Normal
\newcommand{\vnse}{\sigma^2}

\title{Active Learning for Discrete Latent Variable Models}

\author{
  Aditi Jha$^{1,2}$, Zoe C. Ashwood$^{1,3}$, Jonathan W. Pillow$^{1}$
 \\ ~ \\
  $^{1}$Princeton Neuroscience Institute, Princeton University \\
  $^{2}$Dept. of Electrical and Computer Engineering, Princeton University \\
  $^{3}$Dept. of Computer Science, Princeton University \\
  \texttt{\{aditijha, zashwood, pillow\}@princeton.edu}
}

\begin{document}
\maketitle

\begin{abstract} \noindent Active learning seeks to reduce the amount
  of data required to fit the parameters of a model, thus forming an
  important class of techniques in modern machine learning. However,
  past work on active learning has largely overlooked latent variable
  models, which play a vital role in neuroscience, psychology, and a
  variety of other engineering and scientific disciplines. Here we
  address this gap by proposing a novel framework for
  maximum-mutual-information input selection for discrete latent
  variable regression models. We first apply our method to a class of
  models known as ``mixtures of linear regressions'' (MLR). While it
  is well known that active learning confers no advantage for
  linear-Gaussian regression models, we use Fisher information to show
  analytically that active learning can nevertheless achieve large
  gains for \textit{mixtures} of such models, and we
  validate this improvement using both simulations and real-world
  data. We then consider a powerful class of temporally structured
  latent variable models given by a Hidden Markov Model (HMM) with generalized
  linear model (GLM) observations, which has recently
  been used to identify discrete states from animal decision-making data. We show that our method substantially reduces the amount of data needed to fit GLM-HMM, and outperforms a variety of approximate methods based on variational and amortized inference. Infomax learning for latent variable models thus offers a powerful for characterizing temporally structured
  latent states, with a wide variety of applications in neuroscience
  and beyond.
\end{abstract}

\section{Introduction}
Obtaining labeled data is a key challenge in many scientific and machine learning applications. Active learning provides a solution to this problem, allowing researchers to identify the most informative data points and thereby minimize the number of examples needed to fit a model. Bayesian active learning, also known as optimal or adaptive experimental design \cite{verdiKadane92,chaloner_bayesian_1995,Cohn1996statistical,Ryan_Drovandi_McGree_Pettitt_2016}, has had a major impact on a variety of disciplines, including neuroscience \cite{lewi_efficient_2007,Lewi2009sequential,Lewi11,DiMattina11,Gollisch12,shababo_bayesian_2013,DiMattina13,Kim14,ParkM14,Pillow16adaptive}, psychology \cite{Watson83,myung_tutorial_2013,DiMattina2015psych,Watson17,Bak18adaptive},
genomics \cite{Steinke07} 
%and astronomy \cite{loredo_bayesian_2004}.
and compressed sensing \cite{Seeger08a,Seeger08b,Vasisht2014active}. 

The general setting for Bayesian active learning involves a probabilistic model $P(y 
\mid \vx, \theta)$, in which a parameter vector $\theta$ governs the probabilistic relationship between inputs $\vx$ and labels or outputs $y$.  To improve learning of $\theta$, we wish to select inputs $\{\vx_t\}$ that will allow us to best estimate  $\theta$ from the resulting dataset $\{\vx_i, y_i\}_{i=1}^t$. In standard ``fixed-design'' experiments, the inputs are selected in advance, or drawn randomly from a predetermined distribution. In adaptive or ``closed-loop'' experiments, by contrast, the inputs are selected adaptively during the experiment based on the measurements obtained so far. Bayesian active learning methods provide a framework for optimally selecting these inputs, where optimality is defined by a utility function that characterizes the specific learning objective \cite{mackay_information-based_1992,Cohn1996statistical,Roy01,Pillow16adaptive}. 
%Common choices of utility involve minimizing the prediction error in future $y$ or minimizing uncertainty about parameter vector $\theta$.

Despite a burgeoning literature, the active learning field has devoted relatively little attention to latent variable models  \cite{Cohn1996statistical,Hefang2000mixture,Anderson2005HMMs}. Latent variable models (LVMs) represent a class of highly expressive models with a vast range of applications. 
In neuroscience in particular, they have provided powerful descriptions of both neural population activity \cite{Rainer2000,Kemere2008,Miller2010,BYu09,ChenZ09b,escola_hidden_2011,Linderman16jnm,Glaser2020,Zoltowski2020icml,Jha_Morais_Pillow_2021} and animal behavior \cite{Wiltschko15,calhoun_unsupervised_2019,ashwood_mice_2021,bolkan_stone,Weilnhammer2021,Zucchini2008}. 

The key feature of latent-variable-based regression models is that the
relationship between input $\vx$ and output $y$ is mediated by an
unobserved or hidden state variable $z$.  This provides such models
with the flexibility to describe internal states of the system that
cannot be observed directly. However, this flexibility comes with a
cost: the likelihood (and by extension, the posterior) in
LVMs is usually not available in closed-form. This complicates
posterior inference and the calculation of expected utility, both of
which are required for Bayesian active learning algorithms.

To address this gap in the literature, we introduce a Bayesian active learning framework for discrete latent variable models. We develop methods based on both MCMC sampling and variational inference to efficiently %methods to sample from the joint posterior distributions over latents and parameters, and show that these samples can be used to 
compute information gain
and select informative inputs in adaptive experiments.
%\cite{Roy01} 
%which is utilized by most Bayesian active learning frameworks
%, is no longer available in closed form. 
%
We illustrate our framework with applications to two specific families of latent variable models: (1) a mixture of linear regressions (MLR) model; and (2) input-output Hidden Markov Models with generalized linear model (GLM) observations (GLM-HMM). We compare the efficiency of different methods, including a recent method based on amortized inference using deep networks \cite{foster_deep_2021}, and show that in both model families our approach provides dramatic speedups in learning over previous methods. 

\section{Related work}
Bayesian active learning methods have been developed for a wide range of different models, from generalized linear models \cite{Chaloner1984optimal,paninski_asymptotic_2005,Khuri2006glmdesign,lewi_efficient_2007,Lewi2009sequential,Lewi11,bak_adaptive_2016,Bak18adaptive} to neural networks \cite{Cohn1996statistical,DiMattina11,DiMattina13,Cowley17adept,gal_deep_2017,Kirsch2019batchbald,Wu_Niu_Chinazzi_Vespignani_Ma_Yu_2021}. One body of work has focused on Bayesian active learning for models with implicit likelihoods \cite{kleinegesse_bayesian_2020,ivanova_implicit_2021}. Another recent line of work has focused on general-purpose real-time active learning using amortized inference in deep neural networks, an approach known as Deep Adaptive Design (DAD) \cite{foster_deep_2021}.
%, provides a. DAD involves training a neural network, which can then be deployed to output the optimal inputs at each trial during an experiment.
However, the literature on active learning for latent variable models is sparse, limited to a few specific model classes and tasks such as density modeling \cite{Cohn1996statistical,Hefang2000mixture} and state estimation for standard HMMs \cite{Anderson2005HMMs}. 
The approach we develop here grows out of previous work on Bayesian active learning methods for generalized linear models  \cite{lewi_efficient_2007,Lewi2009sequential,houlsby_bayesian_2011,Bak18adaptive}. However, our contribution is novel as we tailor Bayesian active learning for latent variable models, especially those used in neuroscience. 
% because the posterior approximations used in prior work are inapplicable to the latent variable model setting. 

\section{Discrete latent variable models (LVMs)}
Before turning to the problem of active learning, we provide a brief
description of discrete latent variable regression models.  The model
has two basic components:  a prior over the latent
variable and a conditional distribution of the
response given the stimulus and latent.  Formally, this model
architecture can be expressed by a pair of equations:
\begin{align}
 z \; &\sim \; P(z \mid \theta) \\ 
 y \mid \vx, z \; &\sim P(y \mid \vx, z, \theta),
\end{align}
where $z \in \{1, ..., K\}$ is a discrete latent variable governing
the internal state of the system, $\vx \in \mathbb{R}^{D}$ is the
input or stimulus, $y \in \mathcal{Y}$ is the response (which may be
continuous or discrete), and $\theta\in \Omega$ denotes a set of model
parameters governing both prior and conditional response
distributions. Fig.~\ref{fig:activelearningschematic}A shows an illustration of an example discrete latent variable model, where the conditional distribution of the response given the stimulus and latent is given by a generalized linear model. The discrete latent variable $z$ governs which of the three generalized linear models determines the response for a given trial. 

Finally, in a latent variable model, the conditional probability of the
response given the stimulus requires marginalizing over the latent
variable:
%$P(y \mid \vx) = \sum_z P(y \mid \vx, z) P(z)$. 
 \begin{equation}
   \label{eq:cond}
   P(y \mid \vx, \theta) = \sum_{k = 1}^K  P(y \mid \vx, z=k, \theta)
   P(z = k \mid \theta).
 \end{equation}

\section{Infomax learning}
% \section{Active Learning framework}
% \section{Infomax input selection for LVMs}
The general goal of active learning is to select inputs that will allow us to
infer the model parameters $\theta$ in as few trials as
possible. Bayesian active learning formalizes this in terms of a
utility function that specifies the goal of learning, e.g., to
maximize mutual information \cite{Lewi2009sequential}, minimize mean-squared error \cite{Kuck2006}, or 
minimize prediction error \cite{Cohn1996statistical,Roy01}.
 
Here we select as utility the mutual information between response $y$
and the model parameters $\theta$ conditioned on the input
$\vx$. Intuitively, this corresponds to selecting the stimulus for
which the resulting response will provide the greatest reduction in
uncertainty about the model parameters, quantified in bits.  Active
learning with mutual information as utility is commonly known as {\it
 infomax learning}, and it has been widely applied in both machine
learning and neuroscience settings
\cite{mackay_information-based_1992,lewi_efficient_2007,Lewi2009sequential,ParkM14,houlsby_bayesian_2011,Pillow16adaptive,Bak18adaptive,DiMattina11}.

Typical frameworks for infomax learning involve a ``greedy'' approach, where inputs are selected one-at-a-time to maximize information provided by $y$ about $\theta$ on each trial.
In this setting, the experimenter selects the stimulus $\vx_t$ on
trial $t$ according to:
 \begin{equation}\label{eq:infomaxDef}
\vx_{t} = \arg \max_{\vx} I(\theta, y_t \mid \vx, \Dat_{t-1}), 
\end{equation}
where $I$ represents mutual information, $y_t$ is the (as yet unobserved) response on trial $t$, and we
have also conditioned on $\Dat_{t-1} =\{(\vx_\tau,y_\tau)\}_{\tau=1}^{t-1}$, the stimulus-response collected previously in the experiment.  This selection rule is equivalent to
saying that we maximize the expected information gain about $\theta$,
or minimize the expected entropy of the posterior over $\theta$
\cite{mackay_information-based_1992}.

 The mutual information (also known as Shannon information) between $y_t$ and $\theta$ given $\vx$ and $\Dat_{t-1}$, can be
 written in several equivalent forms \cite{CoverThomas}, one of which is: 
\begin{align}
  I(\theta, y_t \mid \vx, \Dat_{t-1}) &= H(y_t \; ; \; \vx, \Dat_{t-1}) -  H(y_t \mid \theta  \;;\;  \vx, \Dat_{t-1}) 
  \label{eq:I1} 
%                        &=  H(\theta  \mid  y \; ; \; \vx) - H(\theta), 
%                        \label{eq:I2} 
%   &= H(y \mid \vx) + H(\theta \mid \vx) - H(\theta, y \mid \vx), 
\end{align}
where 
\begin{align}
\label{eq:condent}
%H(y_t \mid  \theta \; ; \;  \vx, \Dat_{t-1} ) = -\mathbb{E}_{\theta \sim P(\theta \mid \Dat_{t-1})} \left[\int_{\mathcal{Y}} P(y_t \mid \theta, \vx, \Dat_{t-1}) \log P(y_t \mid \theta,\vx, \Dat_{t-1})\, dy_t\right]
H(y_t \mid  \theta \; ; \;  \vx, \Dat_{t-1} ) &= -\int_{\Omega} \int_{\mathcal{Y}} P(y_t,\theta \mid \theta, \vx, \Dat_{t-1}) \log P(y_t, \mid \theta,\vx, \Dat_{t-1})\, dy_t\, d\theta  \\
\intertext{denotes the conditional entropy of $y$ given $\theta$, and}
% Here, $\theta$ is integrated over its posterior distribution $p(\theta \mid \Dat_{t-1})$. 
%Next, 
   \label{eq:margent}
H(y_t \; ; \;  \vx, \Dat_{t-1}) &= -\int_{\mathcal{Y}} P(y_t \mid \vx, \Dat_{t-1}) \log P(y_t \mid \vx, \Dat_{t-1})\, dy_t
\end{align}
is the marginal entropy of $y$, with both terms conditioned on the stimulus $\vx$ and previously collected data $\Dat_{t-1}$.
 % Note that this latter entropy uses the marginal distribution of $y$ given $\vx$ and $\Dat_{t-1}$, which requires an integral over the parameters:
% %\begin{equation}
% $P(y \mid \vx, \Dat_{t-1}) = \int_\Omega P(y,\theta \mid \vx, \Dat_{t-1})\, d\theta$, which may be challenging to compute in some settings.
In the above expressions, the integrals over $y$ can be replaced by sums when $y$ is discrete.

%= \int_\Omega P(y \mid \theta, \vx) P(\theta)\, d\theta$
%$P(y \mid \vx)$, the marginal distribution of $y$ given $\vx$, with the parameters integrated out, that is:

%Here we will focus on the first of these formulations of the mutual information (eq.~\ref{eq:I1}), although other work (e.g., \citeA{Lewi11} has focused on the second (eq.~\ref{eq:I2}).  (See \citeA{Bak18adaptive} for a comparison in the context of Bernoulli generalized linear models).

%This second formulation has the convenient property that because the parameters $\theta$ are generally independent of the stimulus, so $H(\theta \mid \vx) = H(\theta)$, which is simply the entropy of $\theta$ prior to observing the response.  However, the conditional entropy $H(\theta \; ; \; y \mid \vx)$ is 

\section{Infomax learning for discrete LVMs}
%Infomax learning for latent variable models is complicated by the fact that the conditional response distribution, $P(y \mid \vx, \theta)$,  is not available in closed form, but requires marginalization over the latent variable (eq.~\ref{eq:cond}). 

The challenge in applying infomax learning to latent variable models
is that the posterior over the model parameters, $p(\theta \mid \Dat_{t-1})$, as well as the conditional response distribution, $p(y_t \mid \theta, \vx, \Dat_{t-1})$, are not available in closed form, due to the fact that they require marginalization over the latent variable. In fact, for discrete latent variable models, these distributions are not even guaranteed to be unimodal (unlike in generalized linear regression models).
%This makes it hard to compute the conditional entropy of $y$ given $\theta$ (eq.~\ref{eq:condent}). 
Furthermore, the marginal response distribution $P(y_t \mid \vx, \Dat_{t-1})$ in eq.~\ref{eq:margent}, requires marginalizing the conditional response distribution over the parameters:
% over both the latent variable and the parameters:
%   \begin{equation}
%     \label{eq:marg}
%     P(y \mid \vx) = \int \sum_{k = 1}^K P(y \mid z = k, \theta)\, P(z = k
%     \mid \theta)\, P(\theta)\, d\theta,
% \end{equation}
  \begin{equation}
    \label{eq:marg}
    P(y_t \mid \vx, \Dat_{t-1}) = \int  P(y_t \mid \theta, \vx, \Dat_{t-1})\, P(\theta \mid \Dat_{t-1})\, d\theta,
\end{equation}
which exacerbates the problem of rapidly computing and optimizing the mutual information between trials.
 
 To overcome this challenge, we develop two different approaches for infomax active learning in discrete latent variable models: one based on
 sampling \cite{houlsby_bayesian_2011,Bak18adaptive} and another based on variational inference (VI)
 \cite{Blei_2016_VI}, which we describe in the next two sections.

%\begin{SCfigure}
%\includegraphics[width=0.5\textwidth, trim={0 0mm 0 0}]{Figs/ActiveLearningSchematic1_JP.pdf} 

\begin{figure}[t]
    \centering
    \includegraphics[width=\textwidth]{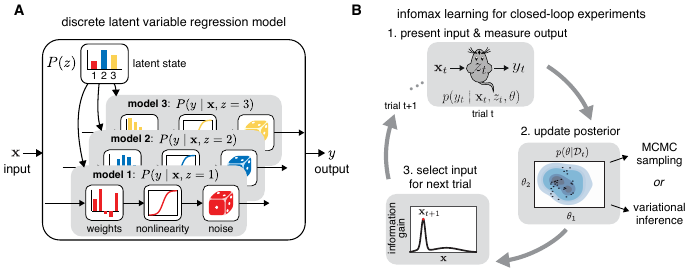} 
    \caption{Discrete latent variable regression models and infomax learning.  \textbf{(A)} Schematic of a discrete latent variable model for regression settings. The response $y$ of the model given a stimulus $x$ and a latent $z$ is produced by generalized linear models. 
Here the discrete latent variable $z$ determines which of the three generalized linear models at the bottom determines the input-output mapping on any trial.  \textbf{(B)} Infomax learning for discrete latent variable models.   On trial $t$, present an input $\mathbf{x}_{t}$ to the system of interest (e.g., a mouse performing a decision-making task) and record its response $y_{t}$. We assume this response depends on the stimulus (input) as well as an internal or latent state $z_{t}$, as specified by the model $P(y_t \mid \vx_t, z_t, \theta)$.  % We want to learn a latent variable model $P(y_t\mid \vx_t, \theta)$, which characterizes the relationship between $\vx_t$ and $y_t$, using parameters $\theta$.
  Second, update the posterior distribution over model parameters $\theta$ given the data collected so far in the experiment, $\Dat_{t} = \{\vx_{1:t}$, $y_{1:t}\}$ using either MCMC sampling or variational inference. Third, select the input for the next trial that maximizes information gain, or the mutual information between the next response $y_{t+1}$ and the model parameters $\theta$.}
    \label{fig:activelearningschematic}
\end{figure}
%\end{SCfigure}

% \begin{figure}[!h]
%     \centering
%     \includegraphics[width=0.7\linewidth, trim = 30mm 100mm 30mm 80mm]{Figs/ActiveLearningSchematic1_AJ.pdf}
%     \caption{Infomax learning for discrete latent variable models: at trial $t$, we present an input $\mathbf{x}_{t}$ to the system we wish to model (for instance, choice behavior in mice \cite{ashwood_mice_2021}) and record its response, $y_{t}$, which depends on the system's latent state $z_{t}$. We want to learn a latent variable model $P(y_t\mid \vx_t, \theta)$, which characterizes the relationship between $\vx_t$ and $y_t$, using parameters $\theta$. In the next step, the posterior distribution over model parameters, $\theta$, is updated to incorporate the data from the most recent trial $t$. Finally, we select the next input to maximize the mutual information between the model's output and the model parameters.}
%     \label{fig:activelearningschematic}
% \end{figure}

\subsection{Sampling-based approach} 
First, we propose a method for infomax learning of discrete latent variable models that relies on Markov Chain Monte Carlo (MCMC) sampling.  Specifically, we use Gibbs sampling to draw samples of $\theta$ from $P(\theta \mid \Dat_{t-1})$, the posterior distribution over parameters given the data collected so far in the experiment. These samples are then used to evaluate the conditional mutual information gain, as described below.

Gibbs sampling allows us to obtain an alternating chain of samples of the latents $z_{1:t-1}$ and the model paramater $\theta$ from their joint conditional distribution  $P(\theta, z_{1:t-1} \mid \vx, \Dat_{t-1})$. As a result, we can get $M$ samples of the model parameter from its posterior at a given trial $t$, $\{\theta^j\}_{j=1}^M \sim P(\theta \mid \Dat_{t-1})$ (effectively marginalizing over the latents). This, however, is not trivial for models where the conditional $P(\theta \mid z_{1:t}, \Dat_{t-1})$ is not available in closed form (such as GLM-HMMs). We developed a modified version of Gibbs sampling for such cases, which we discuss in detail later in sec.~\ref{sec:infomaxforiohmms}.

Each sample, $\theta^j$, obtained using Gibbs sampling parameterizes a model with conditional probability of the response $y$ given by $P(y \mid \theta^{j}, \vx, \Dat_{t-1})$, which can be evaluated by marginalizing over the discrete latents (using eq.~\ref{eq:cond}). This, then, allows us to compute the marginal likelihood of the response $y$:
\begin{align}
\label{eq:sample_marginal}
    P(y \mid \vx, \Dat_{t-1}) \approx \frac{1}{M}\sum_{j=1}^M P(y \mid \theta^j, \vx, \Dat_{t-1})
\end{align}
Using the conditional and marginal likelihoods, we can next compute sample-based versions of the entropy terms (eqs.~\ref{eq:condent} and ~\ref{eq:margent}) as follows:
\begin{align}
\label{eq:sample_entropy}
    H(y \mid \theta\; ; \;  \vx, \Dat_{t-1}) &\approx \frac{1}{M} \sum_{j=1}^M \int_{\mathcal{Y}}  P(y \mid \theta^j, \vx, \Dat_{t-1}) \log  P(y \mid \theta^j, \vx, \Dat_{t-1}) dy \\
    \intertext{and}
    H(y \; ; \;  \vx, \Dat_{t-1}) &= \int P(y \mid \vx, \Dat_{t-1}) \log  P(y \mid \vx, \Dat_{t-1}) dy\\
    \approx \int_{\mathcal{Y}} \Bigg(\frac{1}{M}\sum_{j=1}^M &P(y \mid \theta^j, \vx, \Dat_{t-1})\Bigg) \log \left(\frac{1}{M}\sum_{j=1}^M P(y \mid \theta^j, \vx, \Dat_{t-1})\right) dy.
\end{align}
Substituting the above equations into the expression for mutual information (eq.~\ref{eq:I1}), we obtain a convenient form for the mutual information that we use in our experiments:
\begin{equation}
\label{eq:infomaxkl}
     I(\theta\; ; \;  y \mid \vx, \Dat_{t-1}) \approx \frac{1}{M} \sum_{j=1}^M D_{KL} \left( P(y \mid \theta^j,\vx,\Dat_{t-1}) \mid \mid P(y \mid \vx,\Dat_{t-1}) \right)
\end{equation}
where  $D_{KL}$ is the Kullback-Leibler divergence (KL divergence, a measure of how different one probability distribution is from another, when both are defined on the same sample space). Here:
\begin{multline}
    \qquad D_{KL}\left( P(y \mid \theta^j,\vx,\Dat_{t-1}) \mid \mid P(y \mid \vx,\Dat_{t-1}) \right) \\ = \int_{\mathcal{Y}} P(y \mid \theta^j,\vx,\Dat_{t-1}) \log \frac{P(y \mid \theta^j,\vx,\Dat_{t-1})}{P(y \mid \vx,\Dat_{t-1})}. \qquad 
\end{multline}
%\approx \frac{1}{M} \sum_{j=1}^M P(y \mid \theta^j,\vx)$
In all our experiments, $y \in \mathbb{R}$, so we discretize $y$ allowing us to replace the integrals over $y$ in the above expressions with summations. 

% The predictive distribution itself can be computed by marginalizing over the latents: $P(y \mid \theta^j,\vx) = \sum_{k=1}^K P(z = k \mid \theta^j,\vx) P(y \mid z=k, \theta^j,\vx)$.

Eq.~\ref{eq:infomaxkl} makes clear that information-based active learning can be equivalently seen as comparing the prediction of the models
given by each of the $M$ samples, $P(y \mid \theta^j,\vx, \Dat_{t-1})$, with the
average model prediction $P(y \mid x, \Dat_{t-1})$, and choosing the input
which maximizes the average difference between predictions of
individual models and the consensus model. This shows that infomax
learning can also be seen as a form of ``query-by-committee''
\cite{Settles2009active}.
This sample-based formulation of infomax learning is also referred to as Bayesian Active Learning by Disagreement \cite{houlsby_bayesian_2011,gal_deep_2017}. 

\subsection{Variational approach} 

While Gibbs sampling allows us to accurately draw samples of the model parameter $\theta$ from its posterior $P(\theta \mid \Dat_{t-1})$, it is often slow and computationally inefficient. As an alternative, we therefore explored the use of variational inference (VI) \cite{Blei_2016_VI} to compute a computationally efficient approximation to the posterior distribution over the model parameters $P(\theta \mid \mathcal{D}_{t-1})$. VI is typically faster than Gibbs sampling, but may be less accurate as it requires use of a simplified approximation to the posterior distribution over model parameters.

Here we use mean-field variational inference,  which assumes that the model parameters and the latents are independent of each other:
\begin{equation}
    \label{eq:mean_field_vi}
    q(\theta, z_{1:{t-1}}) = q_1(\theta)q_2(z_{1:t-1})
\end{equation}
where $q_1$ and $q_2$ represent the approximate variational posteriors over $\theta$ and $z_{1:t-1}$ respectively. We first assume simple tractable distributions to be the prior distributions over $\theta$ (such as a multivariate Gaussian) and over the latents (such as an independent categorical distribution for $z$ at every trial). We then use coordinate ascent to optimize the parameters of these assumed distributions in order to minimize the Kullback-Leibler divergence between the approximate and true posteriors:
\begin{equation}
    q^*_1(\theta)q^*_2(z_{1:t-1}) = \arg\min_{q^*_1(\theta)q^*_2(z_{1:t-1})} D_{KL}\left(q^*_1(\theta)q^*_2(z_{1:t-1}) \mid\mid P(\theta, z_{1:t-1} \mid \mathcal{D}_{t-1})\right)
\end{equation}
We describe the coordinate ascent update steps in detail for the model classes that we consider in the appendix (sec.~\ref{sec:vi_iohmms} and sec.~\ref{sec:vi_mlrs}). 

However, having an approximate posterior over $\theta$ is insufficient to compute mutual information in closed form in the setting of discrete LVMs. The conditional response distribution $p(y_t \mid \vx, \Dat_{t-1})$ is still not available in closed form, and is required to compute the conditional entropy of $y$ given $\theta$, as well as the marginal entropy of $y$. Hence, we instead draw samples of the model parameter $\{\theta^j\}_{j=1}^M$ from $q^*_1(\theta)$ (as opposed to the true posterior in case of Gibbs sampling, which makes VI much faster) and then use these samples to compute mutual information as described above in eq.~\ref{eq:infomaxkl}. 

\vspace{1em}
 To summarize, Figure~\ref{fig:activelearningschematic}B shows an illustration of
 infomax active learning for discrete LVMs in the context of a
 neuroscience experiment. The animal receives an input
 $\vx_t$ and generates a response $y_t$ on each trial $t$. In the
 sampling-based approach, we then use samples of the joint
 distribution over latents $z$ and parameters $\theta$ to evaluate the
 expectations required for computing mutual information. In the variational inference-based approach, we compute an approximate posterior over $\theta$ given $\Dat_t$ and then draw samples from it to evaluate mutual information, as given in eq.~\ref{eq:infomaxkl}. Finally,
 we select the stimulus for trial $t+1$ which maximizes the
 conditional mutual information  between the response and model parameters,  $I(y_{t+1}, \theta \mid \vx, \Dat_t)$.

\section{Mixture of linear regressions (MLR)}
\label{sec:infomaxformlrs}
We now illustrate the power of our proposed infomax learning frameworks with applications to specific latent variable models, the first of which is a {\it mixture of linear regressions} (MLR) model.  This model has a rich history in machine learning \cite{li_learning_2018,gaffney_trajectory_1999,bengio_input_1995}.It consists of an independent mixture of $K$ distinct linear-Gaussian regression models (Fig.~\ref{fig:results_mlr}A).
Given an input, $\vx \in \mathbb{R}^D$, the corresponding output observation $y \in \mathbb{R}$ arises from one of the $K$ components as determined by the latent state $z \in \{1,..K\}$. Formally, the model can be described as:
\begin{align} 
    z_t &\sim\text{Cat}(\pi) \\ \quad y_t &\mid \left( \vx_t, z_t = k \right) \sim \Nrm(\vx_t\trp \vw_k, \sigma^2)
    \label{eq:MLR} 
\end{align} 
where $\pi \in \Delta^{K-1}$ denotes a discrete or categorical distribution over the set of $K$ mixing components, and $\vw_k \in \mathbb{R}^D$ denotes the weights of the linear regression model in state $k$. The model parameters to be learned are thus given by $\theta = \{\vw_{1:K}, \pi\}$.   

\subsection{Fisher information analysis}
Before applying our algorithm to the MLR model, it is worth asking whether there is any hope that infomax learning will be helpful in this setting. In the standard linear-Gaussian regression model, it is straightforward to see that posterior covariance of model parameters, given by $(C_0\inv + \frac{1}{\sigma^2} \sum_{t=1}^T \vx_t\vx_t\trp )\inv$ where $C_0$ is the prior covariance, is independent of the outputs $\{y_t\}$. This means that an optimal design can be planned out prior to the experiment, and there is no benefit to taking into account the output $y_t$ when selecting the next input $\vx_{t+1}$  \cite{Chaloner1984optimal,mackay_information-based_1992}. Adaptive experimental design thus provides no benefit for the standard linear-Gaussian regression model. Intriguingly, however, this does {\it not} hold for the MLR model. 

To quantify the asymptotic performance of infomax learning for the MLR model, and to gain insight into which inputs are most informative, we can examine the Fisher information of the MLR model \cite{paninski_asymptotic_2005}. The Fisher information matrix for a model with parameters $\theta$ is a matrix with $i,j$'th element  $J_{ij} = \E \left[\left(\frac{\partial}{\partial \theta_i} \log P(y \mid \vx, \theta) \right)\left(\frac{\partial}{\partial \theta_j} \log P(y \mid \vx, \theta) \right) \right]$, where expectation is taken with respect to $P(y \mid \vx, \theta)$. For an MLR model in $D$ dimensions with $K$ components, the Fisher information matrix for the weights given an input vector $\vx$ is a $KD \times KD$ matrix whose $i,j$'th block is given by:
\begin{equation}
    J_{[i,j]}(\vx) = \tfrac{1}{\sigma^4} \E \big[  (y-\vx\trp \vw_i)(y-\vx\trp \vw_j) P(z = i \mid y, \vx, \theta) P(z=j \mid y, \vx, \theta) \big] \,\vx \vx\trp ,
\end{equation}
where expectation is taken with respect to the marginal response distribution (see Appendix \ref{sec:fisher_mlrs} for details). Although this expectation cannot generally be computed analytically \cite{Behboodian1972information}, we can compute it for two extremal cases of interest: (1) perfect identifiability, when the response $y$ gives perfect information about the latent variable;
%, that is, about which set of linear weights generated the response; 
and (2) perfect non-identifiability, when the response provides no information about the latent variable.  

\begin{figure}
    \centering
    \includegraphics[width=0.9\linewidth]{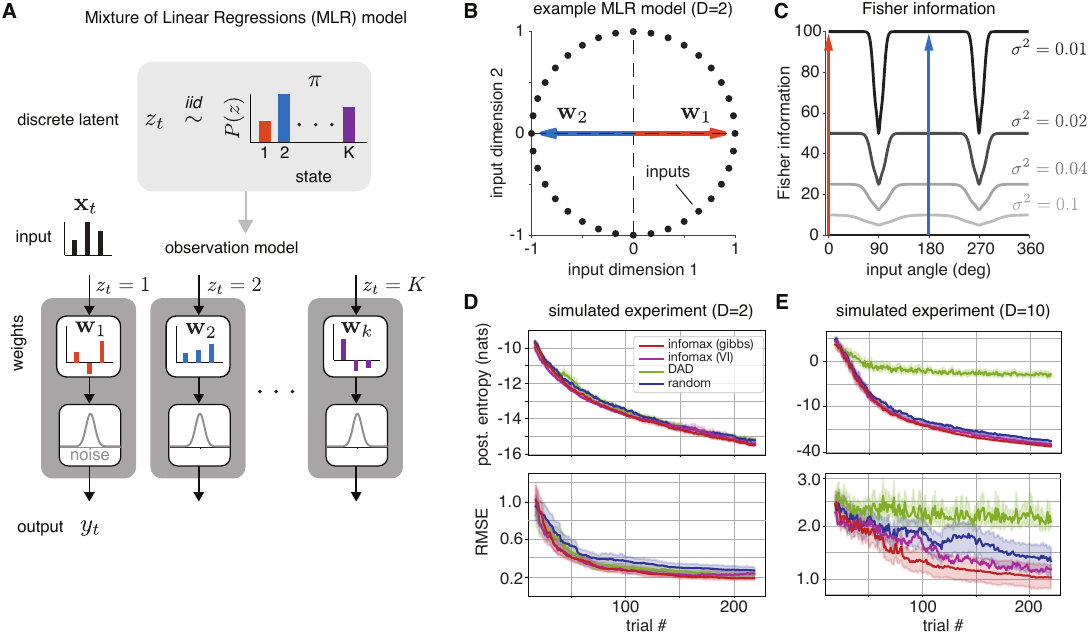}
     \vspace{-0.25 cm}
    \caption{Infomax learning for mixture of linear regressions (MLR) models. \textbf{(A)} Model schematic. At time step $t$, the system is in state $z_{t} = k$ with probability $\pi_{k}$. The system generates output $y_{t}$ using state-dependent weights $\mathbf{w}_{k}$ and independent additive Gaussian noise (Eq.~\ref{eq:MLR}). \textbf{(B)} Example 2-state model with two-dimensional weights $\mathbf{w}_{1} = (1,\,0)$ and $\mathbf{w}_{2} = (-1,\, 0)$. We consider possible inputs on the unit circle, which are the information-maximizing inputs for linear Gaussian models under an $L_2$ norm constraint. (C) Fisher information as a function of the angle between $\mathbf{w}_{1}$ and the input presented to the system, for different noise variances $\sigma^{2}$. 
    (D) Comparison between infomax active learning (using Gibbs sampling and VI), DAD and random sampling for the 2D MLR model shown above with mixing probabilities $\pi = [0.6, 0.4]$ and noise variance $\sigma^2=0.1$. 
    %Posterior entropy and the Root Mean Squared Error (RMSE) between the generative and recovered parameters are plotted as a function of the number of trials of the experiment for both random sampling (blue) and our infomax procedure (red). 
    Error bars reflect 95\% confidence interval (standard error) of the mean across 20 experiments. (E) Performance comparison for the same 2-state model but with 10-dimensional weight vectors and inputs. The possible inputs to the system were uniform samples from the 10-D unit hyper-sphere. }
    \vspace{-1 em}
    \label{fig:results_mlr}
\end{figure}

To illustrate these two cases, Fig.~\ref{fig:results_mlr}B shows an example MLR model with two 2D weight vectors pointing in opposite directions along the $x_1$ axis.  If observation noise variance $\sigma^2$ is small, a unit vector input with 0 degree orientation renders the latent state perfectly identifiable, since the response will be large and positive if $z=1$ and large and negative if $z=2$.  On the other hand, an input at 90 or 270 degrees gives rise to perfect non-identifiability; these inputs are orthogonal to both $\vw_1$ and $\vw_2$, so observing the output $y$ will provide no information about which model component (weights $\vw_1$ or $\vw_2$) produced it. 

In the case of perfect identifiability, the Fisher information matrix simplifies to a block diagonal matrix with $\frac{1}{\sigma^2} \pi_i \vx \vx\trp$ in its $i$th block (see \ref{sec:fisher_mlrs}).  The trace of the Fisher information matrix, which quantifies the total Fisher information provided by this input, is $\frac{1}{\sigma^2} ||\vx||^2$, which remarkably, is the same Fisher information as in the standard (non-mixture) linear regression model.  In the case of non-identifiability, on the other hand, the Fisher information is a rank-1 matrix with block $i,j$ given by $\frac{1}{\sigma^2} \pi_i \pi_j \vx\vx\trp$.  In the case where all class prior probabilites are equal ($\pi_i = 1/K \ \forall i$), the trace is only $\frac{1}{K\sigma^2} ||\vx||^2$, revealing that non-identifiable inputs can provide as little as $1/K$ as much Fisher information as inputs with perfect identifiability. The dependence on the number of components, $K$, is worth noting as it suggests that active learning yields larger improvements  for models with more components.

Figure \ref{fig:results_mlr}C shows the (numerically computed) Fisher information as a function of input angle for the MLR model shown in panel B, for different noise levels $\sigma^2$. This confirms the analytic result that Fisher information for this 2-state MLR model is 1/2 its maximal value for inputs in the ``non-identifiable'' region, and shows that this sub-optimal region grows wider as noise variance increases. %Our analysis suggests even larger improvement in models with more than two states. 
This analysis confirms that active learning can improve MLR model fitting, and shows that the most informative inputs are those that (in addition to have large $L_2$ norm) provide information about the discrete latent variable.

%However, it applies only to the asymptotic regime and cannot be used to select stimuli on a trial-by-trial basis.

\subsection{Infomax learning algorithm for MLR}
To perform infomax learning for MLR models, we used Gibbs sampling \cite{bishop_pattern_2006} to obtain samples from the posterior over model parameters. This involved sampling from the joint distribution over the latents $z$ and the model parameters $\theta = \{\vw_{1:K},\pi\}$, conditioned on the data. As an alternate strategy, we also drew samples of the model parameters from a variational approximation to its posterior
%input-output pair from the current time step, $(\vx_{t}, y_{t})$, as well the input-output pairs from all past time steps, $\Dat_{t-1}$.
(see \ref{sec:gibbs_sampling_mlrs} and \ref{sec:vi_mlrs} for details). Next, using $M$ samples of the model parameters, $\{\vw_{1:K}^j, \pi^j\}_{j=1}^M$, we computed the mutual information between the system's output and the parameters  (Eq.~\ref{eq:infomaxkl}) for a grid of candidate inputs by substituting the likelihood term, $P(y \mid \theta^j, \vx, \Dat_t) = \sum_{k=1}^K \pi^j_k \mathcal{N}(y \mid \vw_k^j \cdot \vx, 1)$, into Eq.~\ref{eq:infomaxkl}. Finally, we selected the input $\vx$ that maximized Eq.~\ref{eq:infomaxkl} and presented it to the system on the next trial. 

%The infomax input-selection procedure is then repeated either for a fixed number of iterations, or until a problem-specific error metric (such as the posterior entropy that we discuss below) has converged.

\begin{figure}[!h]
    \centering
    \includegraphics[width=0.8\linewidth]{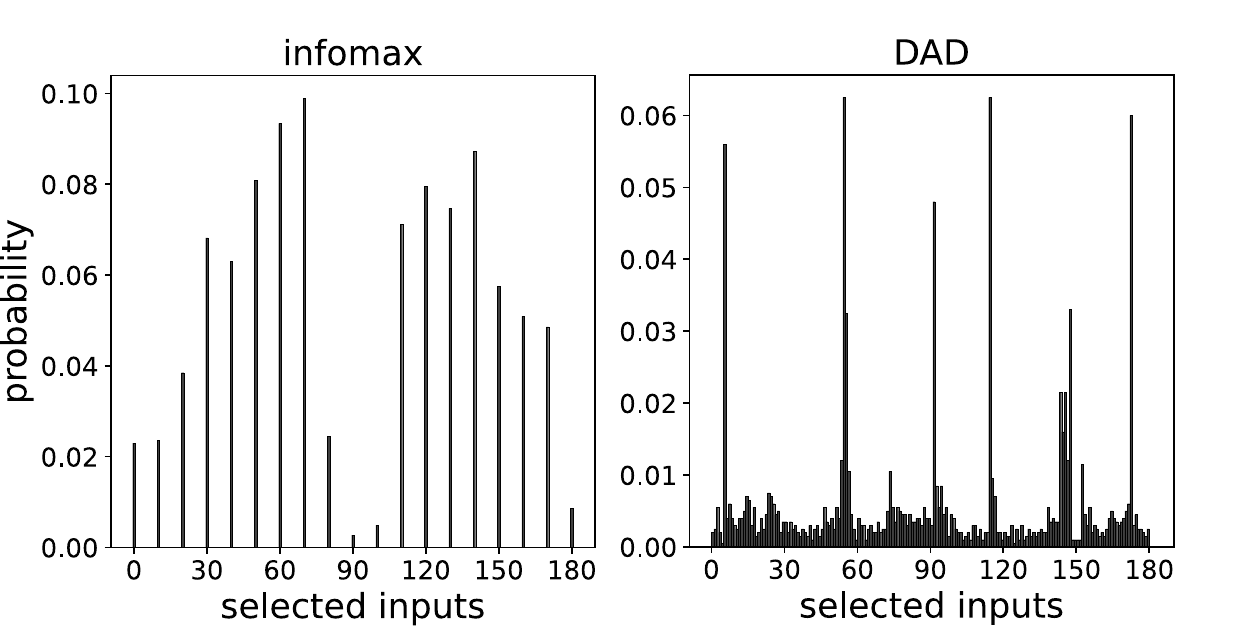}
    \caption{Histogram showing inputs selected by our active learning method (over the course of 200 trials) on mixture of linear regressions (MLRs), when inputs lie on a 2D circle  (see Fig. \ref{fig:results_mlr}). We find a drop in probability at 90$^\circ$, this is also predicted by the Fisher Information analysis discussed in text for infomax (using Gibbs sampling). However, we do not see such a trend while using DAD. Inputs selected by DAD were distributed over the unit circle with modes at multiple of 30$^\circ$. (DAD requires a continuous range of inputs, hence it select inputs from all over the unit circle as opposed to a discrete list.)}
    \label{fig:inputs_selected}
\end{figure}

\subsection{Numerical experiments for MLRs}
\label{sec:mlrexps}
To evaluate our active learning framework for MLRs, we first performed two simulations. In our first experiment, illustrated in Fig.~\ref{fig:results_mlr}B, we considered a grid of possible inputs on the unit circle, spaced $10^{\circ}$ apart. (This was motivated by the fact that the optimal stimuli for the linear regression model have maximal $L_2$ norm, and thus lie on the surface of a hypersphere centered at zero).  On every trial, we selected an input from this set and sampled the output from one of $K=2$ regression models. We fixed the state probabilities as $\pi = [0.6, 0.4]$. The regression models had the form: $y_t = w_k\trp x_t + \epsilon$ where we fixed the generative parameters as: $\vw_1 = [-1,0]$, $\vw_2 = [1,0]$ and $\epsilon \sim \mathcal{N}(0,0.1)$. 

Our second experiment followed the same setup, but selected inputs from a set of $1000$ candidate points sampled uniformly on the 10-D hyper-sphere. The output again arse from one of the two regression models, now with weights oriented along the first two major axes, $\vw_1 = [1,0,...,0]$ and $\vw_2 = [0,1,0...0]$, again with mixing weights $\pi = [0.6, 0.4]$.

The task at hand is to learn the generative parameters of the model: $\{\vw_1, \vw_2, \pi\}$. We compared several input-selection strategies including our infomax learning methods (using Gibbs sampling and variational inference), a random sampling approach which selected inputs uniformly from the set of all possible inputs, and the Deep Adaptive Design (DAD) method proposed by \cite{foster_deep_2021}. We adapted the code for DAD to use it for input selection in MLRs (details in \ref{sec:training_dad}). In all cases, after input selection, at each trial we inferred the model parameters using Gibbs sampling. 

A natural quantity to track during infomax learning is the entropy of the posterior distribution over the model parameters $\theta$ \cite{Bak18adaptive}, which we approximate as $\log(|\text{cov}(\theta)|)$ (we drop the additional term $\frac{D}{2}(1+2\pi)$ as it is constant for our experiments). We computed the sample estimate of this posterior entropy using the $M=500$ samples obtained from Gibbs sampling at every trial. We found that posterior entropy decreased fastest for infomax with Gibbs sampling (``Gibbs-infomax'', top panel of Fig.~\ref{fig:results_mlr}D). In 10-d, this difference was even more prominent (top panel of Fig.~\ref{fig:results_mlr}E). 
We also tracked the root mean squared error (RMSE) between the true and estimated parameters. The bottom panel of Fig.~\ref{fig:results_mlr}D shows that for the 2D simulation, RMSE decreased fastest for Gibbs-infomax stimulus selection. 

Finally, Figure ~\ref{fig:results_mlr}E shows that in a model with 10D inputs, RMSE decreased fastest for Gibbs-infomax, followed by infomax with variational inference (``VI-infomax''). This shows that evaluating information gain using samples from the true posterior produced substantially better learning than with samples from the variational posterior. Furthermore, while DAD was comparable to VI-infomax learning 2 dimensions, it did not perform well for high-dimensional inputs. We feel these results were particularly impressive given that RMSE was \textit{not} the objective function we optimized. Overall, our proposed Gibb-infomax algorithm produced highly sample-efficient learning of MLRs in comparison to other methods.

In case of 2D inputs, this improvement can be attributed to the fact that  Fisher Information drops dramatically when the angle between the weight vectors and the input is close to $90^{\circ}$ or $270^{\circ}$ (as discussed above). Hence, our active learning strategy outperformed random sampling by avoiding the uninformative inputs orthogonal to the model weights.  Fig.\ref{fig:inputs_selected} shows that our algorithm did indeed avoid these inputs. As the Fisher information analysis given above makes clear, higher dimensionality leads to increased probability that randomly selected inputs will fall in the region of non-identifiability (i.e., be orthogonal to all of the model weight vectors $\vw_k$), given that random vectors in high dimensions have high probability of being orthogonal \cite{gorban_blessing_2018}. This aligns with our finding that the benefits of active learning are more pronounced in higher dimensions.

\subsection{Application: CA Housing Dataset}
\label{sec:mlrhousing}
\begin{figure}[h]
    \centering
    \includegraphics[width=\linewidth]{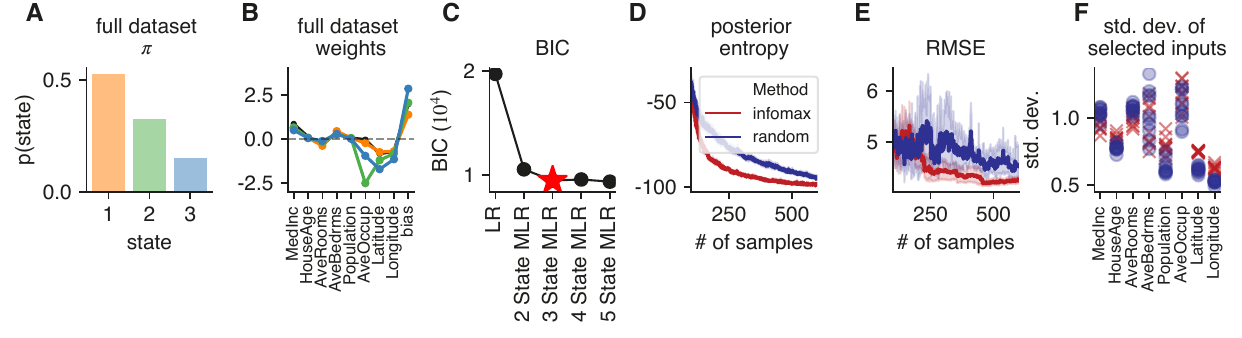}
    \vspace{-2em}
    \caption{Application of infomax learning to CA Housing Dataset \cite{kelley_pace_sparse_1997}. \textbf{(A)} Best fitting mixing weights for 3 state MLR to 5000 samples of the dataset. \textbf{(B)} Best fitting state weights for 3 state MLR to 5000 samples of the CA housing dataset. Orange, green and blue represent states 1, 2 and 3 respectively. Black represents the linear regression fit. \textbf{(C)} BIC as number of MLR states is varied from 1 (standard linear regression) to 5. We select the 3 state model as BIC begins to level off beyond 3 states.  \textbf{(D)} Posterior entropy between the 3 state MLR parameters obtained using 5000 samples (parameters shown in (A) and (B)) and recovered parameters as a function of the number of samples for random sampling (blue) and infomax with gibbs sampling (red). 
    Error bars reflect 95\% confidence interval of the mean across 10 experiments. \textbf{(E)} The same as in (D) but for the RMSE (root mean squared error).  \textbf{(F)} Visualization of standard deviation of 500 inputs selected by both infomax (red) and random sampling (blue). Each dot corresponds to a different experiment.  Examining (B), it is clear that the 3 states differ most according to the weights placed on the `AveOccup', `Latitude' and `Longitude' covariates. All 10 infomax experiments select inputs with greater variance for the latitude and longitude covariates than are selected by the random sampling experiments.}
    \label{fig:results_housing}
\end{figure}

Finally, we applied infomax learning to the California housing dataset of \cite{kelley_pace_sparse_1997}. This dataset contains the median house price, in 1990, as well as 8 predictors of house price for 20,640 census block groups. The dataset is accessible via scikit-learn \cite{scikit-learn}. We fit MLRs with different numbers of states to a reduced dataset of 5000 samples 
% (we used the reduced dataset to reduce the time required to fit the MLRs) 
and found that a 3 state MLR described the CA housing dataset well (Figure ~\ref{fig:results_housing}C), and offered a dramatic improvement in predictive power relative to standard linear regression (a 1 state MLR).  Figures \ref{fig:results_housing}A and \ref{fig:results_housing}B show the best fitting mixing weights and state weights for this 3 state MLR. Next, we wanted to understand if infomax learning would allow us to learn the best-fitting 3 state MLR parameters with fewer samples. Figures \ref{fig:results_housing}D and \ref{fig:results_housing}E show that Gibbs-infomax learning did indeed substantially reduce the number of samples required to learn the model parameters.

In Figure \ref{fig:results_housing}B, it is clear that the three discrete states differed  most according to the weights placed on the `AveOccup' (average occupancy), `Latitude' and `Longitude' covariates.  Intriguingly, in Figure \ref{fig:results_housing}F, we see that the inputs selected by infomax learning had greater variance for the Latitude and Longitude covariates compared to those selected with random sampling (the red crosses are always above the blue dots).  This is a useful external validation that infomax selects inputs in a manner that accords with intuition.

\section{Input-Output Hidden Markov Models (IO-HMM)}
\label{sec:infomaxforiohmms}

The Input-Output Hidden Markov Model (IO-HMM) represents a powerful extension of the standard HMM \cite{bengio_input_1995}. A standard HMM has $K$ discrete states, a fixed transition matrix that describes the probability of transitions between states, and a distribution over outputs for each state.  Crucially, at each time step $t$, the observed output $y_t$ in a standard HMM depends only on the current state, $z_t \in \{1,..K\}$.   IO-HMMs have an additional component: an external input vector presented at every time step: $\vx_t \in \mathbb{R}^D$. As a result, both state transitions and observations can depend on the input vector.

Recent work in neuroscience has focused on a class of IO-HMMs in which the input-output mapping is parametrized by a generalized linear model (GLM), resulting in a model known as the GLM-HMM \cite{escola_hidden_2011,calhoun_unsupervised_2019,ashwood_mice_2021,bolkan_stone}.  Here, we consider the Bernoulli GLM-HMM, which assumes that the outputs are binary, $y_t \in \{0,1\}$, and are produced according to state-specific GLM weights, $\vw_k \in \mathbb{R}^D$:
\begin{align}
    \label{eq:bernoulliGLM}
        P(y_t = 1 \mid \vx_t, z_t=k) =\frac{1}{1+\exp^{-\vw_{k}\trp \vx_t}}
\end{align}
We assume that, as in the standard HMM, state transitions are governed by a stationary, input-independent transition matrix, $A \in \mathbb{R}^{K\times K}$, where $A_{il} = P(z_t=l \mid z_{t-1} = i)$. The first state $z_1$ has prior distribution $\pi \in \Delta^{K-1}$.  The GLM-HMM model parameters are thus $\theta = \{ \vw_{1:K}, A, \pi\}$.

To perform infomax learning for GLM-HMMs, we use Gibbs sampling to iteratively sample the latent states $\{z_1, \ldots z_t\}$ for all trials observed so far given the model parameters $\theta$, and the model parameters $\theta$ given the sampled latents after each trial (step 2 in Fig.~\ref{fig:activelearningschematic}). Gibbs sampling-based inference for HMMs is well-known \cite{Ghaharamni2001HMM}. However, when we use Bernoulli-GLM observations, the conditional distribution over $\{\vw_{1:K}\}$ is no longer available in closed form since there is no conjugate prior distribution for the weights of a Bernoulli GLM. Thus, we developed a method for sampling  $\{\vw_{1:K}\}$ using Laplace approximation (see Appendix  \ref{sec:gibbs_iohmm} for details). An alternative strategy for sampling from logistic models involves using Polya-Gamma augmentation \cite{Polson2013pg,pillow_fully_2012}. We compared these two approaches and found that our Laplace-based approach performed equally well to Polya-Gamma augmentation (see \ref{sec:additional_iohmms}), thus empirically validating our method.

For comparison, we also developed an approximate infomax learning algorithm using variational inference (VI). We used mean-field VI to obtain posterior distributions over the model parameters $\theta$. Because there is no conjugate prior for the GLM weights $\{\vw_{1:K}\}$, we used the Laplace approximation to approximate their posteriors (see \ref{sec:vi_iohmms}). After updating the variational posterior distribution on each time step, we drew samples of model parameters from their variational posteriors in order to evaluate the information gain associated with each candidate stimulus.

During infomax learning with GLM-HMMs (step 3 of Fig.~\ref{fig:activelearningschematic}), we used $M=500$ samples of the model parameters, $\{\vw_{1:K}^j, A^j, \pi^j\}_{j=1}^M$, to compute the mutual information between the output and the model parameters according to Eq.~\ref{eq:infomaxkl}. Here, the likelihood for the GLM-HMM is:
\begin{equation}
    P(y \mid \theta^j, x, \Dat_t) = \sum_{k=1}^K P(z = k \mid \Dat_t, \theta^j) P(y \mid x, z = k)
\end{equation}
where $P(z = k \mid \Dat_t,\theta^j)$ can readily be obtained using the forward-backward algorithm and $P(y \mid x, z=k)$ is the Bernoulli-GLM likelihood function (Eq.~\ref{eq:bernoulliGLM}). We computed the mutual information over a discrete set of candidate inputs and the selected the most informative input to present on the subsequent trial.
% = \sum_{l=1}^K A^j_{lk} P(z_{t}=l \mid \Dat_t, \theta^j)$
% \begin{figure}
%     \centering
%     \includegraphics[width=\linewidth,trim =40mm 20mm 40mm 20mm, clip=True]{Figs/Fig4GLMHMM.pdf}
%     \caption{Results; describe each panel}
%     \label{fig:glmhmm1}
% \end{figure}

\subsection{Numerical experiments with GLM-HMM}
\label{sec:iohmmsexps}
\begin{figure}[!h]
    \vspace{-1em}
    \centering
    \includegraphics[width=0.87\linewidth]{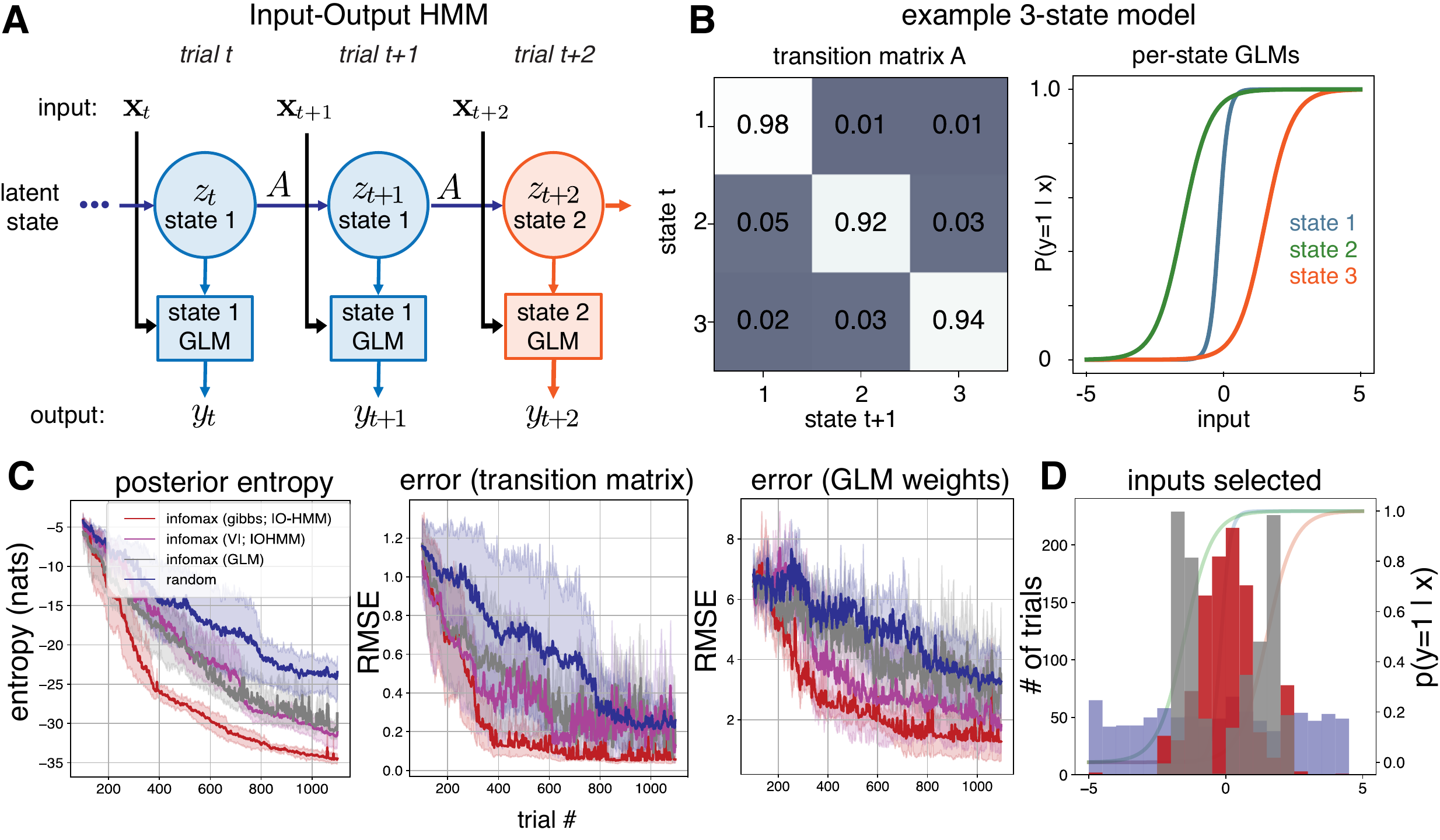}
     \vspace{-0.3cm}
    \caption{Infomax for GLM-HMMs. (\textbf{A}) Data generation process for the GLM-HMM. At time step $t$, a system generates output $y_{t}$ based on its input, $\mathbf{x}_{t}$, as well as its latent state at that time step, $z_{t}$. The system then either remains in the same state, or transitions into a new state at trial $t+1$, with the transition probabilities given by matrix $A$. (\textbf{B}) Example settings for the transition matrix and state GLMs for a 3 state GLM-HMM. These are the settings we use to generate output data for the analyses shown in panels C and D. (\textbf{C}) Left: posterior entropy over the course of 1000 trials for random sampling (blue), infomax with a single GLM (grey), infomax for the full GLM-HMM using variational inference (VI) and Gibbs sampling (magenta and red respectively).  Middle: root mean squared error for the recovered transition matrix for each of the three input-selection schemes (random/infomax with GLM/infomax with GLM-HMM (Gibbs)/infomax with GLM-HMM (VI)). Right: root mean squared error for the weight vectors of the GLM-HMM for each of the input-selection schemes. (\textbf{D}) Selected inputs for random sampling (blue), active learning when there is model mismatch and the model used for infomax is a single GLM (gray), active learning with infomax (using Gibbs sampling) and the full GLM-HMM (red). Selected inputs over the course of 1000 trials are plotted, and are shown on top of the generative GLM curves. }
    \vspace{-1em}
    \label{fig:glmhmm}
\end{figure}
We sampled data from the 3-state GLM-HMM (Fig.~\ref{fig:glmhmm}B). We set the model parameters to closely approximate those inferred from mice performing a binary sensory decision-making task  \cite{ashwood_mice_2021}. Each GLM has a weight ($w_k$) associated with the external stimulus as well as a bias parameter ($b_k$), such that the GLM weight vector is $\vw_k = \{w_k, b_k\}$. The input stimuli ($x_t$) are thus 1-dimensional, such that the choice probability can be formally written as:
\begin{equation}\label{eq:BernoulliGLM}
    P(y_t = 1 \mid \vx_t, z_t=k) = \frac{1}{1+\exp^{-w_k x_t + b_k}}
\end{equation}
In our experiment, we selected inputs from a grid of stimuli over the range $[-5,5]$, spaced 0.01 units apart. Similar to the MLR setting, the task here is to recover the true parameters of the GLM-HMM used to simulate data. We compared the performance of our infomax learning methods (based on either Gibbs sampling or variational inference) as well as a ``random sampling'' approach in which inputs were sampled uniformly at random.  Deep Adaptive Design (DAD) \cite{foster_deep_2021} is not applicable in this setting as it assumes trials to be i.i.d., and thus we did not consider it.
% In both cases, we use $500$ samples of the model parameters from Gibbs sampling (after discarding the first $200$ samples as burn-in). 

We examined the performance of these three methods, and found that the posterior entropy over the model parameters decreased  fastest under Gibbs-infomax, followed by VI-infomax, and was slowest with random sampling (Fig.~\ref{fig:glmhmm}C, left). We also observed that the RMSE between the true and inferred parameters decreased much faster for our active learning methods (with best performance under Gibbs-infomax) as compared to random sampling, both for the transition matrix $A$ (middle panel of Fig.~\ref{fig:glmhmm}C) and the GLM weights (right panel of Fig.~\ref{fig:glmhmm}C). This suggests that our infomax learning method can be used to fit GLM-HMMs using fewer samples. It also reinforces our previous result that sampling from the exact posterior substantially benefits infomax learning as compared to using the variational posterior.

To understand why our framework outperforms random sampling for the GLM-HMM, we plotted histograms of the inputs selected by random sampling and by Gibbs-infomax (Fig.~\ref{fig:glmhmm}D). While random sampling selected inputs from the entire input domain, infomax learning rarely selected inputs with a magnitude greater than 3. For positive inputs $>3$, the sigmoid nonlinearity (Eq.~\ref{eq:BernoulliGLM}) saturated for all three models, so that sampled $y_{t}$ are 1 with high probability and are thus uninformative about the latent state. Similarly, for large-magnitude negative inputs, the $y_{t}$ samples are 0 with high probability for all three states. As such, the outputs generated by these provide virtually no information about the latents (necessary for updating the transition matrix) or the GLM weights. Overall, infomax learning substantially reduced the number of samples required to learn the parameters of the GLM-HMM.

To make our method practical for closed-loop experiments, it is critical for it to compute new inputs quickly. For example, in the case of mouse decision-making experiments, consecutive trials occur within 1--10 seconds \cite{pinto_accumulation--evidence_2018,laboratory_standardized_2020-1}. While our current implementation requires up to 20s per trial (on an 1.7GHz quad-core i7 laptop), we show in the appendix (\ref{sec:additional_iohmms}) that running five parallel chains of 100 samples each provides a $5$x speedup over the current implementation with a single chain of 500 Gibbs samples. Additionally, we also performed infomax learning for a special case of GLM-HMMs: mixtures of GLMs. We show in \ref{sec:infomax_mglms} that our method outperforms random sampling in terms of posterior entropy and error in recovering the model parameters. These results provide further evidence that our infomax learning method is applicable across model settings. 

\subsection{Consequences of ignoring latent states}
% Active learning with a specific model could result in selecting inputs that are biased towards that model. 
To assess the importance of latent structure on active learning methods, we benchmarked our method against an additional input-selection scheme: infomax under conditions of model mismatch. Specifically, we compared to a strategy where inputs were selected by infomax under the (mismatched) assumption that responses arose from a single Bernoulli-GLM, with no latent states. %This involves generating outputs from the true GLM-HMM when it is presented with inputs selected by infomax assuming a Bernoulli-GLM as the underlying model of the system. 
This allowed us to explore the effect of ignoring the presence of latent variables when selecting inputs.

Fig.~\ref{fig:glmhmm}D shows that the inputs selected by Bernoulli-GLM infomax learning differed substantially from those selected by the full GLM-HMM infomax algorithm. In particular, the Bernoulli-GLM method avoided selecting inputs in both the center and the outer edges of the input domain. In virtue of neglecting the outer edges, it outperformed random input selection (compare the grey and blue lines in all panels of Fig.~\ref{fig:glmhmm}C). However, the full GLM-HMM infomax method still performed best for learning the weights and transition matrix of the true model (red lines in Fig.~\ref{fig:glmhmm}C). The significant drop in the performance when ignoring the presence of latent states thus highlights the importance of developing active learning methods tailored specifically for latent variable models.  %Our active learning method with models that do not incorporate latent states performs poorly as compared to using a latent variable model. This emphasizes the importance of developing active learning methods for latent variable models. 

\subsection{Downstream application: latent state inference}
\begin{SCfigure}
    \centering
    \includegraphics[width=0.45\columnwidth]{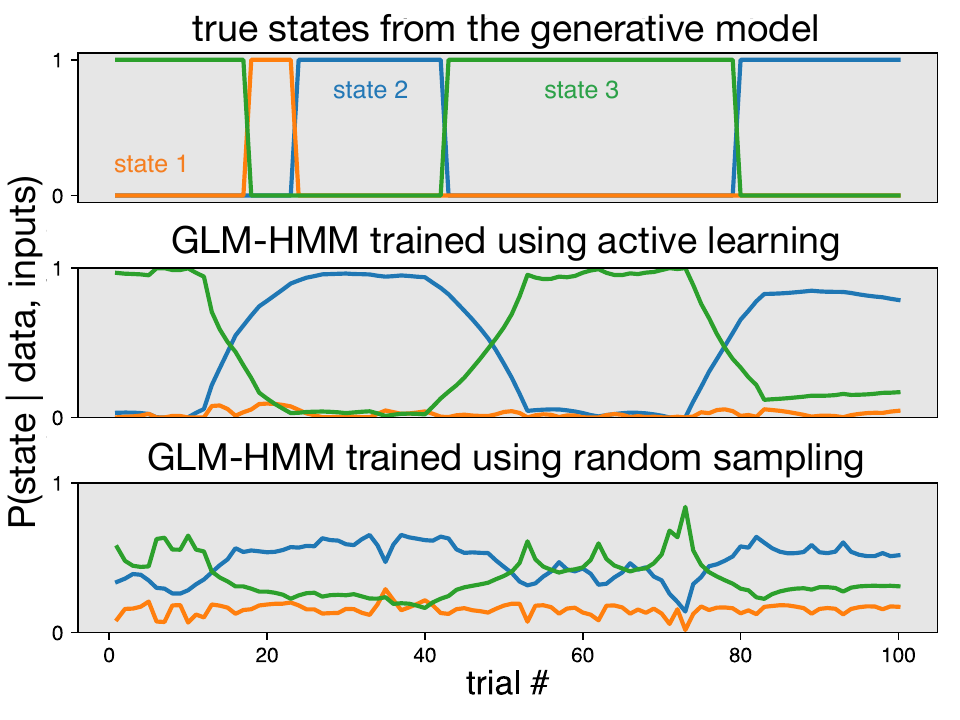}
    % \vspace{-0.5cm}
    \caption{Inferring latent states. (Top) the true latent states of the data-generating GLM-HMM for 100 trials. (Middle) the inferred posterior probabilities of states using an GLM-HMM, trained using infomax learning on 400 trials from the data-generating GLM-HMM. (Bottom) the same for an GLM-HMM trained using random sampling on 400 trials from the data-generating GLM-HMM.}
    \label{fig:state_prediction}
\end{SCfigure}

GLM-HMMs are often used to infer the underlying latent states during the course of an experiment. To demonstrate the utility of our active learning approach for downstream tasks, we compare infomax learning and random sampling for predicting latent states across trials. We use the same generative GLM-HMM as shown in Fig.~\ref{fig:glmhmm}B, and train two new distinct GLM-HMMs using 400 input-output samples from the generative model. One of the GLM-HMMs is trained using inputs selected by infomax learning (with Gibbs sampling), while the other is trained using random input selection. Next, we generate a set of 100 trials from the generative model, and use the two GLM-HMMs to predict the posterior probabilities of states at each trial. Fig.~\ref{fig:state_prediction} shows that the GLM-HMM trained using infomax learning is able to predict the true states drastically better than that trained using random selection using the same number of trials. 

\subsection{Infomax for Mixture of GLMs (MGLMs)}\label{sec:infomax_mglms}
\begin{figure*}[!h]
    \centering
    \includegraphics[width=1.0\linewidth, trim = 10mm 60mm 10mm 40mm, clip=True]{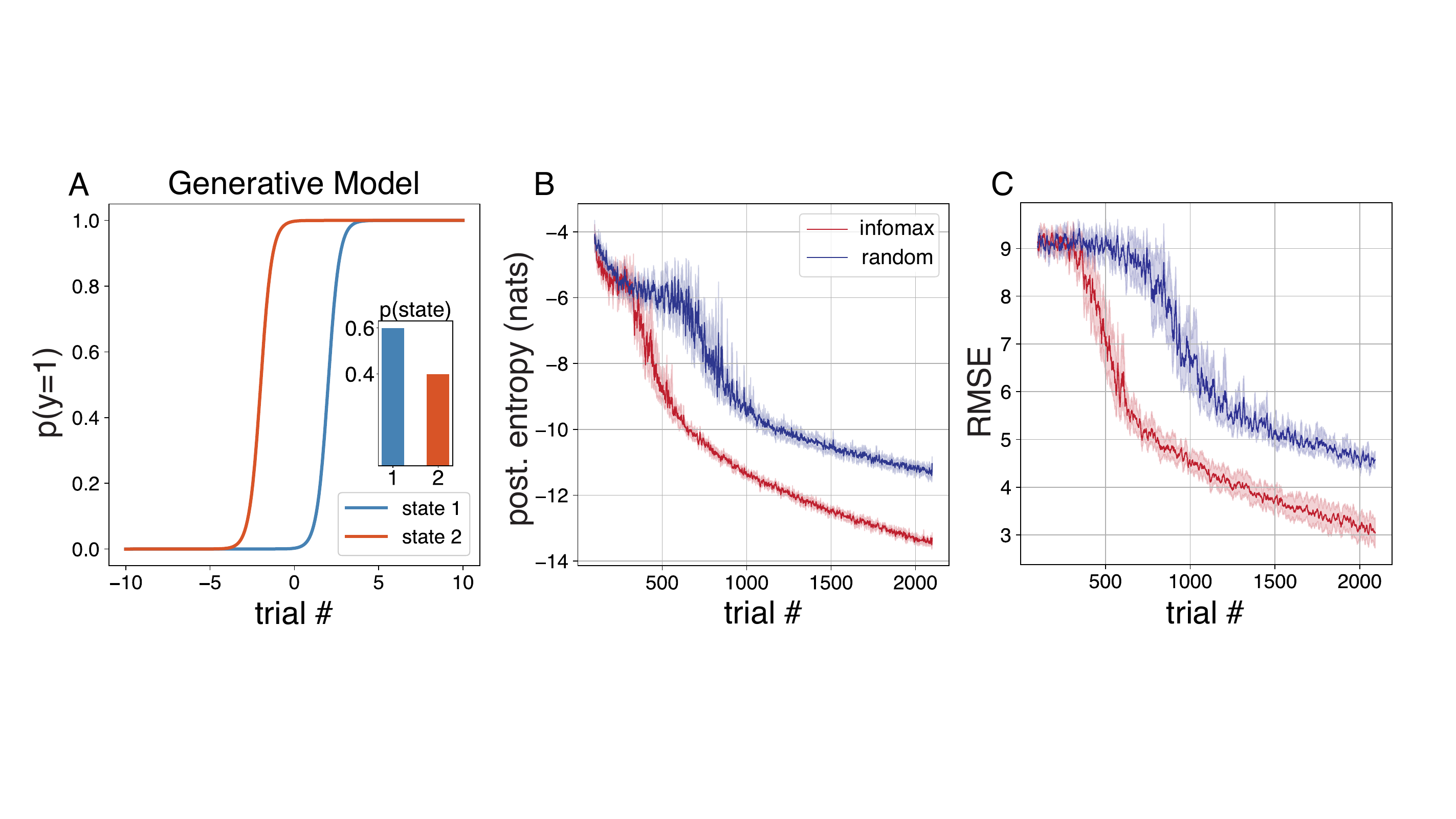}
    \caption{Infomax learning for mixture of GLMs (MGLMs): (A) Data generation model. Example settings for a 2-state MGLM along with the mixing weights for the two states. 
    %These settings are used for the experiments shown in panels B and C. 
    (B) Posterior entropy of model parameters over the course of 2000 trials for random sampling (blue) and infomax learning for MGLM (blue). (C) Root mean squared error for the recovered GLM weights and mixing weights for each of the two input-selection schemes.}
    \label{fig:moglm}
\end{figure*}

Finally, we evaluate infomax on a special case of GLM-HMMs: a mixture of Bernoulli-GLMs (MGLMs). Compared to standard GLM-HMMs, MGLMs assume that the probability that the system transitions to state $k$ at trial $t+1$ is independent of the system's state at trial $t$. MGLMs arise in a number of settings including in medicine, transport modeling and in marketing \cite{farewell_use_1988,follmann_generalizing_1989,follmann_identifiability_1991,wedel_mixture_1995,li_application_2018}. Formally, MGLMs contain $K$ distinct GLM observation models where the state of the model, $z\in\{1,...K\}$, is independently sampled at each time step from a distribution $\pi \in \Delta^{K-1}$. Similar to the GLM-HMM setup, observations are generated according to a Bernoulli GLM as in Eq.~\ref{eq:bernoulliGLM}. Infomax learning using Gibbs sampling for MGLMs involves similar steps to those required for GLM-HMMs and is described in \ref{sec:gibbs_moglm}.
%entails using Gibbs sampling to iteratively sample the latent states, $z_{t'} \ \forall \ t'\in[1,t]$ and the model parameters $\{\vw_{1:K}, \pi\}$. While the latent states at each time step are sampled independently without the need for the forward-backward algorithm, we use the same procedure as in GLM-HMMs to sample the model parameters $\{\vw_{1:K}, \pi\}$.

We perform an experiment to assess the effectiveness of our active learning method in this setting. Data was generated from a 2-state MGLM model (shown in Fig.~\ref{fig:moglm}A) with $\pi = [0.6, 0.4]$ and the GLM weights $w_1 = [3,-6], w_2 = [3, 6]$. We find that our active learning method is better than random sampling at inferring the parameters of this model (Fig.~\ref{fig:moglm}B, C). 

\section{Discussion}
\label{sec:discussion}
We have developed novel methods for Bayesian active learning in discrete latent variable models (LVMs). We applied these methods to two classes of models: mixture of linear regressions and input-output HMMs. We showed that infomax learning consistently achieved lower error and lower posterior entropy than random input selection. Our method also outperformed  active learning methods that ignored the presence of latent variables and, for the case of MLRs, the DAD method of \cite{foster_deep_2021}.

Given the importance of LVMs in neuroscience \cite{escola_hidden_2011,calhoun_unsupervised_2019,ashwood_mice_2021,bolkan_stone} and other scientific domains, we envisage broad applicability of our method. One exciting application is to adaptively select stimuli in animal decision-making tasks. While recent work has shown that the behavior of mice can be well-described with a multi-state GLM-HMM \cite{ashwood_mice_2021,bolkan_stone}, they required large amounts of data collected from multiple sessions across days. Using our framework, it may be possible to learn these parameters using data from a single day, reducing the time and cost of experiments and thereby speeding up scientific discovery.

Now, we briefly discuss some limitations of our work. First, we considered scalar output observations. Extending to higher-dimensional outputs may require alternate methods for computing information, since numerical integration in high-d is difficult. Second, we selected maximally informative inputs from a discrete set of candidate inputs on each trial. Future work may instead use optimization to find optimal inputs in a continuous input space. A final direction for future work is to consider GLM-HMMs in which state transitions also depend on the input. Despite these limitations, our method substantially speeds up the learning of systems characterized by latent variable models, and will be highly beneficial in neuroscience and other fields with time-consuming or expensive experiments.

% \printbibliography
\bibliographystyle{apalike}
\bibliography{bibliography}
%%%%%%%%%%%%%%%%%%%%%%%%%%%%%%%%%%%%%%%%%%%%%%%%%%%%%%%%%%%%
\appendix
\section{Appendix}

\setcounter{figure}{0}
\setcounter{table}{0}
\setcounter{equation}{0}
\setcounter{section}{0}

\renewcommand{\thefigure}{A\arabic{figure}}
\renewcommand{\thetable}{A\arabic{table}}
\renewcommand{\theequation}{A\arabic{equation}}
\renewcommand{\thesection}{A\arabic{section}}

\section{Gibbs Sampling for MLRs}\label{sec:gibbs_sampling_mlrs}
Here, we describe Gibbs Sampling algorithm for Mixture of Linear Regressions models.
Given $T$ trials, for each input-output pair, $\vx_t \in \mathbb{R}^D$ and $y_t \in \mathbb{R}$, we sample class-belongings, $z_t \in \{1,..K\}$, from:
\begin{equation}
    P(z_t = k \mid y_t, \vx_t, \vw_{1:K}, \pi, \sigma) = \frac{\mathcal{N}(y_t; \vw_k\trp \vx_t, \sigma^2)\pi_k}{\sum_l \mathcal{N}(y_t; \vw_l\trp \vx_t, \sigma^2)\pi_l}
\end{equation}
Next, we sample new estimates of the mixing parameters from:
\begin{equation}
    \pi_k \mid z_{1:T} \sim \text{Dir}(n_k + 1)
\end{equation}
where $n_{k} = \sum_{'t=1}^{T}\mathbb{1}(z_{t'} = k)$. 

Finally, we assume a Gaussian prior, $\Nrm(\vw_0, \sigma_0^2 I)$, over the weights associated with each latent class and sample a new estimate for them as follows:
\begin{align}
    \vw_k &\sim \mathcal{N}(\vw_{k}', \Sigma_{k}')\\
    \vw_{k}' &= \vw_0 +  (\sigma_0^2I + X_k X_k\trp)^{-1} X_k\trp(Y_k - X_k \vw_0)\\
    \Sigma_{k}' &= I -  X_k\trp \left( \sigma_0^2I + X_k X_k\trp \right)^{-1} X_k. 
\end{align}
Here, the rows of $X_k \in T_k \times D$ and $Y_k \in T_k \times 1$ contain inputs and outputs at time points where $z=k$, respectively. We fix $\vw_0 = \vzero$ and $\sigma^2_0 = 10$ in our experiments. 
 We perform this procedure $M$ times in order to obtain $M$ samples of the model parameters, $\{\vw_{1:K}^j, \pi^j\}_{j=1}^M$, where $M = 500$ (excluding 100 burn-in samples) in our experiments. 
 
 \section{Variational inference for MLRs}\label{sec:vi_mlrs}
Here, we describe mean-field variational inference for MLRs, which we use to derive posterior distributions over the model's parameters. Following mean-field approximation, we assume independence between all the model parameters and the latent variables. 

Given $T$ trials, for each input-output pair, $\vx_t \in \mathbb{R}^D$ and $y_t \in \mathbb{R}$, we assume that it's mixture assignment $z_t \in \{1,...K\}$ is governed by an independent categorical distribution $q(z_t; \phi_t)$, where $\phi_t \in \Delta^{K-1}$. We, further, assume that the weight $\vw_k \in \mathbb{R}^D$ of the $k$-th linear regression model has a normal posterior distribution $q(\vw_k; \mu_k, \Sigma_k)$, with mean $\mu_k \in \mathbb{R}^D$ and covariance $\Sigma_k \in \mathbb{R}^{D \times D}$. Hence:
\begin{equation}
    q(\vw_{1:K}, z_{1:T}) = \prod_{t=1}^T q(z_t; \phi_t) \prod_{k=1}^K q(\vw_k; \mu_k, \Sigma_k) 
\end{equation}

Let us vertically stack $\phi_t$ for $t\in {1:T}$ and denote this by a matrix $\boldsymbol{\phi}$ of size $T\times K$. Similarly, let $X \in \mathbb{R}^{T\times D}$ represent the design matrix with all inputs stacked, and $Y \in \mathbb{R}^{T\times 1}$ contain all observations. Also, we know that each of the linear regressions in the MLR model has Gaussian noise with variance $\sigma^2$.

We update the variational parameters $\phi_t$, $\mu_{1:K}$ and $\Sigma_{1:K}$ iteratively using the update rules described below. For each $t \in \{1..T\}$,
\begin{equation}
    \phi_{tk} \propto  \exp \{y_t x_t\trp \mathbb{E}[ \mu_k] - \mathbb{E}[ (x_t\trp \mu_k)^2]/2\}
\end{equation}
Next, for each $k \in \{1...K\}$, we assume a Gaussian prior distribution over the weights: $\Nrm(\vw_0, \sigma_0^2 I)$, we update the variational parameters governing the weights as follows: 
\begin{align}
    \Sigma_k &= \left( \sigma_0^2 I + \frac{1}{\sigma^2} \left(\left(\boldsymbol{\phi}_{:,k} \mathbf{1}\right) \cdot X\right)\trp X \right)^{-1}\\
    \mu_k &= \frac{1}{\sigma^2}\Sigma_k X\trp (\boldsymbol{\phi}_{:,k} Y) 
    \end{align}
We fix $\vw_0 =\boldsymbol{0}$ and $\sigma^2=10$ in our experiments.
We repeat these updates until either the log-likelihood of the data arising from the model has converged or a limit of 500 iterations has reached. 

Once the variational posteriors have been learned, we draw $M$ samples each, for the weights $\vw_{1:K}$ and the mixture assignments $z_{1:T}$. Finally, using the mixture assignments, we obtain $M$ samples for the mixing probability $\pi$ by computing the proportion of trials assigned to each state. We set $M=500$ in our experiments, thus obtaining $\{\vw^j_{1:K}, \pi^j \}_{j=1}^{500}$.

 \section{Training details for Deep Adaptive Design (DAD)}\label{sec:training_dad}
 We downloaded the code for DAD, and adapted it to perform input selection for MLRs (which we attach with the supplement). The parameters of the MLR model were set to the same values as described in sec.~\ref{sec:mlrexps}. Since the DAD model requires continuous inputs, rather than a discrete list of inputs, we allow it to choose inputs from the unit circle in $2$d and the unit hypersphere in $10$d, rather than restricting it to the discrete set of stimuli in sec.~\ref{sec:mlrexps}. 
 
 The DAD model has two components: the encoder network which takes in input-observation pairs $\{x, y\}$ and outputs an encoding for this. This is a feedforward neural network. We set this network to have 3 layers: the input layer which has 3 nodes for the first MLR experiment ($2$d inputs and $1$d observations) and 11 nodes for the second experiment ($10-$d inputs and $1$d observations), a hidden layer with 256 nodes and ReLu activation function, and a linear output layer with 16 nodes.
 
 Following this, the encoded history is taken as input by an emitter network. This network outputs the input for the next trial: $x_{t}$. The input layer of this feedforward network has the same dimensionality as the output of the embedding layer, i.e. 16 nodes. It has one hidden layer with ReLu activation and 256 nodes, followed by a linear output layer with as many nodes as the dimensionality of the input to the MLR model. We normalize the output of this network, to ensure that the selected $x_{t}$ lies on the unit circle/unit hypersphere.
 
 We do a hyperparameter optimization to select the number of hidden layers and nodes from the range of values used in the experiments (no. of hidden layers: 1--3, no. of nodes per layer: 16/128/256) in the original DAD \cite{foster_deep_2021} paper. 
 
To compute the sPCE loss that DAD uses to optimize the two neural networks, we use $500$ samples each to compute the inner and outer expectation in the loss function. Since our experiments involve large number of trials ($T=200$), we use score gradient estimator to compute the gradients that are backpropagated while training. Finally, we train the model using Adam (with betas set to 0.8, 0.998), and use exponential learning rate annealing (where the initial learning rate is set to 1e-4 post a search over the range 1e-5--1e-3, and $\gamma = 0.96$) for a total of 50000 gradient steps. 

% \section{Inputs selected during active learning for MLRs}\label{sec:inputs_mlrs}
% We show the inputs selected by our active learning algorithm as well as by DAD \cite{foster_deep_2021} in Fig.~\ref{fig:input_selection_mlrs}.
% \begin{figure}[h]
%     \centering
%     \includegraphics[width=0.8\linewidth]{Figs/inputsselected.pdf}
%     \caption{Histogram showing inputs selected by our active learning method (over the course of 200 trials) on mixture of linear regressions (MLRs), when inputs lie on a 2D circle  (see Fig. \ref{fig:results_mlr}). We find a drop in probability at 90$^\circ$, this is also predicted by the Fisher Information analysis discussed in text for infomax (using Gibbs sampling). However, we do not see such a trend while using DAD. Inputs here are randomly distributed over the unit circle, with modes at multiple of 30$^\circ$. (DAD requires a continuous range of inputs, hence we let is select inputs from all over the unit circle as opposed to a discrete list, as in Fig. \ref{fig:results_mlr}.)}
%     \label{fig:input_selection_mlrs}
% \end{figure}

\section{Fisher Information for MLRs}\label{sec:fisher_mlrs}

Here we derive the Fisher information for the weights of the MLR model
(shown in Fig.\ref{fig:results_mlr}B of the main text).

We consider a model consisting of a mixture of $K$ linear
regression models in a $D$-dimensional input space, defined by weights
$\{ \vw_1, \vw_2, \ldots, \vw_K \}$.  The full model weights take the
form of a length-$KD$ vector formed by stacking the weights for
each component:
\begin{equation}
  \label{eq:1}
 \vw =
 \begin{bmatrix}
 \vw_1 \\ \vdots \\ \vw_K  
 \end{bmatrix}.
\end{equation}
The Fisher information $J$ is a $KD \times KD$ matrix carrying the
expectation for the product of partial derivatives of the
log-likelihood with respect to each element of $\vw$.  We will 
derive the $D \times D$ blocks of the Fisher information matrix for each pair of
components in $\{1, \ldots, K\}$.

The block of partial derivatives for component $j$ is given by:
\begin{align}
  \label{eq:2}
  \frac{\partial}{\partial \vw_j} \log p (y \mid \vx, \theta)
  &=
    \frac{1}{\vnse} (y - \vx\trp \vw_j) \vx  \left(\frac{ \pi_j
    \exp{\left(-\frac{1}{2 \vnse} (y-\vx\trp \vw_j)^2\right)}}{P(y \mid
    \vx, \theta)} \right) \nonumber \\
 &=
   \frac{1}{\vnse} (y - \vx\trp \vw_j) \vx 
   \left(  \frac{P(y \mid \vx, z=j, \theta) P(z = j \mid \pi)} {P(y \mid \vx,
     \theta)} \right)
   \nonumber \\
  &=
    \frac{1}{\vnse}
    (y - \vx\trp \vw_j) \vx\, \Big(P(z = j \mid  y, \vx,\theta) \Big).
\end{align}
Plugging this into the formula for Fisher information, we obtain the
following expression for the $i,j$'th block of the Fisher information
matrix:
 \begin{equation}
     J_{[i,j]}(\vx) = \frac{1}{\sigma^4}  \E \Big[  (y-\vx\trp
     \vw_i)(y-\vx\trp \vw_j) P(z = i \mid y, \vx, \theta) P(z=j \mid
     y, \vx, \theta) \Big] \,\vx \vx\trp , 
 \end{equation}
where expectation is taken with respect to the marginal distribution
$P(y \mid \vx, \theta)$.  This expectation cannot in general be
computed in closed form (see \cite{Behboodian1972information}).
However, we considered two special cases in the text where an analytic
expression is available.

\subsection{Perfect identifiability}

First, the case of ``perfect identifiabilty'' arises when the
conditional distributions $P(y \mid \vx, z = j, \theta)$ are
well-separated for the different classes of latent variable $z$, or
equivalently, the posterior class probabilities
$P(z = j \mid y, \vx, \theta)$ are effectively 0 or 1 for virtually all output
values $y$.  In practice, this arises for inputs $\vx$ such that the
conditional means $\{\vx\trp \vw_1, \vx\trp \vw_2, \ldots, \vx\trp\vw_K\}$ are well separated relative to
the noise standard deviation $\sigma$ (e.g., more than $2\sigma$
apart).  In this case, the off-diagonal blocks of the Fisher
information matrix are zero, since $P(z=i \mid y, \vx, \theta) P(z=j
\mid y, \vx, \theta) \approx 0 $ for $i \neq j$.  The diagonal
blocks, by contrast, can be computed in closed form:
\begin{align}
     J_{[j,j]}(\vx) &= \frac{1}{\sigma^4}  \E \Big[  (y-\vx\trp
                      \vw_j)^2 P(z = j \mid y, \vx, \theta)^2 \Big]
                      \vx \vx\trp \nonumber
  \\
 &= \frac{1}{\sigma^4}  \left(\int_{-\infty}^\infty  (y-\vx\trp
   \vw_j)^2 \pi_j \Nrm(y \mid \vx\trp w, \vnse)  dy \right)  \vx
   \vx\trp \nonumber
 \\
   &= \frac{1 }{\vnse} \pi_j\vx \vx\trp. 
\end{align}
We can write the Fisher information matrix efficiently as:
\begin{equation}
  \label{eq:4}
  J(\vx) = \frac{1}{\vnse}\diag(\pi) \otimes \vx\vx\trp,
\end{equation}
where $\otimes$ denotes the Kronecker product. The trace of the Fisher
information is
\begin{equation}
  \label{eq:3}
  \Tr[J] = \frac{1}{\vnse} \Tr[\diag(\pi)] \Tr[\vx\vx\trp] =
  \frac{1}{\vnse} \vx\trp\vx,
\end{equation}
which is the trace of the Fisher information matrix in the standard
linear-Gaussian regression model. This confirms---as one might
expect---that in the case of perfect identifiability we have the same
amount of Fisher information as in a model without latent variables.

\subsection{Non-identifiability}

Second, the case of ``non-identifiabilty'' arises when the conditional
distributions $P(y \mid \vx, z = j, \theta)$ are identical for the
different classes of latent variable $z$, meaning the output $y$
carries no information about the mixing component that generated it.
This arises when the linear projection of $\vx$ onto all of the weight
vectors is identical,
$\vx\trp \vw_1 = \vx\trp\vw_2 = \cdots = \vx\trp \vw_K$.  This arises,
for example, when the stimulus is orthogonal to all of the weight
vectors, which occurs with high probability in high-dimensional
settings.  

In this case we can also compute the Fisher information in closed
form. We obtain, for block $i,j$ of the Fisher information matrix:
\begin{align}
     J_{[i,j]}(\vx) &= \frac{1}{\sigma^4}  \E \Big[  (y-\vx\trp
                      \vw_i)^2 \pi_i \pi_j  \Big] \vx \vx\trp
\nonumber  \\
 &= \frac{1}{\sigma^2}  \pi_i \pi_j \vx\vx\trp,
\end{align}
where we have used the fact that $\vx\trp\vw_i = \vx\trp \vw_j$ and
that the product of posterior probabilities
$P(z=i \mid y, \vx,\theta)P(z=j \mid y, \vx,\theta)$ is equal to the product
of prior probabilities $\pi_i \pi_j$ in the setting where the output
$y$ carries no information about the latent $z$.

The Fisher information matrix can be written in Kronecker form:
\begin{equation}
  \label{eq:5}
  J = \frac{1}{\vnse} (\pi \pi\trp) \otimes \vx \vx\trp,
\end{equation}
which has trace
\begin{equation}
  \label{eq:6}
  \Tr[J] = \frac{1}{\vnse} (\pi\trp \pi) \vx\trp \vx.
\end{equation}
This expression is minimal when the prior probabilities are all equal
to $1/K$, in which case $\pi\trp \pi = 1/K$, giving $\Tr[J] =
\frac{1}{K\vnse} \vx\trp \vx$.

\section{Gibbs Sampling For GLM-HMMs} \label{sec:gibbs_iohmm}
We provide a complete description of Gibbs sampling for GLM-HMMs in Alg.~\ref{alg:iohmmgibbs}. It uses outputs $y_{1:T}$ and inputs $\vx_{1:T}$, along with the prior over model parameters to provide $M$ samples of the latent states $\{{z}_{1:T}\}^j$ as well as of the model parameters $\{{\vw}_{1:K}$, $A$, $\pi \}^j$. We assume the model has $K$ distinct latent states. Sampling the latent states (Alg.~\ref{alg:stateseqsample}) requires using backward messages, $B_{t,k} = P(y_{t+1:T} \mid x_{1:T}, z_t=k)$, which can be obtained using standard forward-backward algorithm \cite{bishop_pattern_2006}.  To sample the weights of the GLMs per state, we use the Laplace approximation followed by an acceptance-rejection step detailed in Alg.~\ref{alg:glmsample}.
We fix the Dirichlet prior $\alpha \in \mathbb{R}^{K+1 \times K}$ over the rows of the transition matrix, $A$, and the initial state distribution, $\pi$ to be a matrix of ones. The GLM weights have an identical prior: $\Nrm(\vzero,10)$. Further, we run Gibbs sampling for 500 iterations and discard the first 100. 
\begin{algorithm}[h]
  \caption{GLM-HMM Gibbs Sampling}
  \label{alg:iohmmgibbs}
    \begin{algorithmic}[1]
      \State \textbf{Input}: Observations $y_{1:T}$, Inputs $\vx_{1:T}$, Prior hyperparameters: $\alpha$, $\vw_0$, $\sigma_0$
      \State \textbf{Output}: Samples $\{({z}_{1:T}$, ${\vw}_{1:K}$, $A$, $\pi)^{(j)} \}$
      \Statex
      \State Initialize $z_{1:T}$, $\vw_{1:K}$, $A$, $\pi$
      \For{$j \gets 1,...M$}
        \For{$k \gets 1,...K$}
            \State $\vw^j_k \gets$ \Call{GLMsampleposterior}{\{$y_t, \vx_t \mid z_t=k \}_{1:T}, \vw_{0},\sigma_{0}, \vw^{j-1}_k$}
            \State $A_{k,:}^j \gets \sample \text{Dir}(\alpha_{k,:} + \mathbf{n}_{k,:})$ 
             \Comment{where $n_{kl} = \sum_t \mathbf{I}(z_t=k,z_{t+1}=l)$}
        \EndFor
        \State $z_{1:T}^j \gets $ \Call{IOHMMsamplestate}{$\pi, A, L$} \Comment{s.t. $L_{t,k} = P(y_t \mid \vx_t, \vw_k)$} 
        \State $\pi^j \gets \sample  \text{Dir}(\alpha_{0,:} + \mathbb{I}_{z_1})$
      \EndFor
    \end{algorithmic}
\end{algorithm}

\begin{algorithm}[h]
 \caption{GLM sample weight from posterior}
 \label{alg:glmsample}
\begin{algorithmic}[1]
    \State \textbf{Input: } Observations $y_{1:T'}$, Inputs $x_{1:T'}$, Prior: $\vw_0, \sigma_0$,  Previous estimate of $\vw$: $\vw^{\text{old}}$
    \State \textbf{Output: } $\{\vw\}$
    \Function{GLMsampleposterior}{ ($y_{1:T'}$, $x_{1:T'}, \vw_0, \sigma_0, \vw^{\text{old}}$)}
        \State $L(\vw) = \sum_{t=1}^{T'} \log P(y=y_t \mid x_t, \vw)$
        \State $\vw^{\text{MAP}} \gets \text{argmax}_{\vw} \left(L(\vw) + \log \Nrm(\vw; \vw_0, \sigma_0^{2} I)\right)$
        \State $C \gets - \left( \frac{\partial^2 L(\vw)}{d {\vw}^2} - \sigma_0^{-2} I  \right)^{-1} {\bigm{\vert}}_{\vw_{\text{MAP}}}$
        \State $\vw^{*} \gets \sample \mathcal{N}(\vw^{\text{MAP}}, C)$
        \State $\alpha(\vw^{*}, \vw^{\text{old}}) \gets \min \left( 1, \frac{\Tilde{p}(\vw^{*} \mid y_{1:T'}, x_{1:T'})  \mathcal{N}(\vw^{old}; \vw^{\text{MAP}}, C)}{\Tilde{p}(\vw^{old}\mid y_{1:T'}, x_{1:T'}) \mathcal{N}(\vw^{*}; \vw^{\text{MAP}}, C)}\right)$ \Comment{$\tilde{p}$: unnormalized posterior}
        \If{$\alpha(\vw^{*}, \vw^{\text{old}}) \geq U(0,1)$}
            \State $\vw \gets \vw^{*}$
        \Else
            \State $\vw \gets \vw^{\text{old}}$
        \EndIf
  \EndFunction
\end{algorithmic}
\end{algorithm}

\begin{algorithm}
  \caption{GLM-HMM State sequence sampling}
  \label{alg:stateseqsample}
\begin{algorithmic}
  \State \textbf{Input}:  Initial state dist. $\pi$, Transition matrix $A$, Likelihood matrix $L \in \mathbb{R}^{T \times K}$
  \State \textbf{Output}: $z_{1:T}$
  \Function{IOHMMsamplestate}{($\pi, A, L$) }
  \State $B \gets \text{HMM-Backwardmessages}(A, L)$ \Comment{$B_{t,k} = P(y_{t+1:T} \mid x_{1:T}, z_t=k)$ \cite{bishop_pattern_2006}} 
    \State $z_1 \gets \sample \pi_k B_{1,k} L_{1,k}$ \text{over } $k \in \{1,...K\}$\\
    \For{$t \gets 2,...T$}
      \State $z_t \gets \sample A_{z_{t-1}, k} B_{t,k} L_{t,k}$ \text{over } $k \in \{1,...K\}$
    \EndFor
  \EndFunction
\end{algorithmic}
\end{algorithm}

\newpage
\section{Variational inference for GLM-HMMs}\label{sec:vi_iohmms}
For a GLM-HMM with $K$ distinct states and Bernoulli-GLM observations, we want to learn variational posteriors for the initial state distribution $\pi_0 \in \Delta^{K-1}$, the transition matrix $A \in \mathbb{R}^{K \times K}$ and the weights of the GLMs, $\vw_{1:K} \in \mathbb{R}^{D}$. To do so, we use inputs to the model $\vx_{1:T}$ and their corresponding observations $y_{1:T}$. The unknown latent states corresponding to these trials are represented by $z_{1:T}$. 

Let's first first define prior distributions over the model parameters:
\begin{align}
    \pi_0 &\sim Dir(\boldsymbol{\alpha}_0)\\
    A_{j,:} = \pi_j =&\sim Dir(\boldsymbol{\alpha}_j) \ \ \ j=1...K\\
    \vw_k &\sim \mathcal{N}(\vw_0, \sigma_0^2 I) \ \ \ k=1...K
\end{align}
where $\boldsymbol{\alpha}_0 \in \mathbb{R}^{K}$ and $\boldsymbol{\alpha}_j \in \mathbb{R}^{K}$ and contain positive real numbers only, $\vw_0 \in \mathbb{R}^D, \ \sigma_0\in \mathbb{R}$.
Now, let us define a variational posterior over the parameters and latent states of the GLM-HMM as follows:
\begin{align}
    q(z_{1:T}, A, \pi_0, \phi_{k=1}^K) = q(z_1)\prod_{t=2}^T q(z_t \mid z_{t-1})q(A)q(\pi_0)\prod_{k=1}^K q(\phi_k)
\end{align}
Here, we assume that the latents are independent of the model parameters, which reflects the mean-field assumption. Next, we develop a coordinate ascent algorithm to iteratively learn the variational posteriors.

We will initialize $q(\pi_0), \ q(A), \ q(\vw_k)$ to their prior distributions. Then, in the first step, we compute the following quantities:
\begin{align}
\label{eq:step1_VIIOHMMs}
    \Tilde{\pi}_0 &= \exp\{\E_{q(\pi_0)}[\ln \pi_{0}]\}\\
    \Tilde{A}_{j,:} &= \exp\{\E_{q(A)}[\ln A_{j,:}]\}\\
    \Tilde{L}_{t,k} &= \exp\{\E_{q(\vw_k})[\ln P(y_t \mid \vw_k, \vx_t)]\} = \exp \Big\{\frac{1}{N}\sum_{i=1}^N \ln P(y_t \mid \vw^i_k, \vx_t) \Big\}
\end{align}
The Dirichlet distributions over $\pi_0$ and $A_{j,:}$ provide closed form updates for $\Tilde{\pi}_0$ and $\Tilde{A}_{j,:}$ (in particular, for a $D-$dimensional vector $x\sim Dir(\boldsymbol{\gamma)}, \ \mathbb{E}[\ln x_i] = \psi(\gamma_i) - \psi(\sum_i \gamma_i)$, where $\psi$ is the digamma function).
To compute $\Tilde{L}_{t,k}$, which is not available in closed form in the case of GLM observations, we obtain a sample estimate of the expectations using 10 samples.

Next, using the quantities computed above, we run forward-backward algorithm for GLM-HMMs \cite{bishop_pattern_2006}, and obtain the forward and backward messages $F, \ B \in \mathbb{R}^{T\times K}$.
This leads to the following distributions over the latent states. 
\begin{align}
    q(z_t = k) &= F_{t,k}B_{t,k}/\left(\sum_{k'}B_{T,k'}\right)\\
    q(z_{t-1}=j, z_t=k) &= F_{t-1,j}\tilde{A}_{j,k} \tilde{L}_{t,k}B_{t,k}/\left(\sum_{k'} B_{T,k'}\right)
\end{align}

Now, we are ready to update the variational distributions over the model parameters:
\begin{align}
    q(\pi_0) & \propto \prod_{k=1}^K \pi_{0k}^{\alpha_{0k} + q(z_1=k) -1}\\
    q (A) & \propto \prod_{k=1}^K \pi_{jk}^{\alpha_{jk} + \sum_{t=2}^T q(z_{t-1}=j, z_t=k) -1}
\end{align}
And finally, the variational approximation over the GLM weights is as follows:
\begin{align}
    q (\vw_k) &\propto \exp \Big\{ \sum_{t=1}^T q (z_t=k) \ln P(y_t \mid \vw_k, \vx_t) + \ln P(\vw_k) \Big\}
\end{align}
Unlike typical Gaussian HMMs, this is not available in closed form because the likelihood of a Bernoulli-GLM does not have a conjugate prior. To deal with this, we approximate $q(\vw_k)$ by a Gaussian distribution using Laplace approximation. Let $L(\vw_k) = \exp \Big\{ \sum_{t=1}^T q (z_t=k) \ln P(y_t \mid \vw_k, \vx_t) + \ln P(\vw_k) \Big\}$.
\begin{align}
\label{eq:laplace_VIIOHMMs}
    q(\vw_k) \sim \Nrm(\vw_k', \Sigma_k')
    ; \quad \vw_k' = \text{argmax}_{\vw_k} L(\vw_k), 
    \quad \Sigma_k' = \left(\frac{\partial^2  L(\vw_k)}{\partial \vw_k^2}\right)^{-1} \big{\vert}_{\vw_k'}
\end{align}
We repeat the update equations from eq.~\ref{eq:step1_VIIOHMMs} to eq.~\ref{eq:laplace_VIIOHMMs} iteratively until the log-likehood of the data from the model converges or a maximum of 500 iterations is reached. 

Once we have obtained a variational distribution for all the model parameters, we can draw $M$ samples of $\{\pi^j_0, A, \vw^j_{1:K}\}_{j=1}^M$ from their variational posteriors. We set $M=500$ for our experiments. 

\section{Additional analyses for GLM-HMMs}\label{sec:additional_iohmms}

\begin{figure*}[!h]
    \centering
    \includegraphics[width=0.8\linewidth]{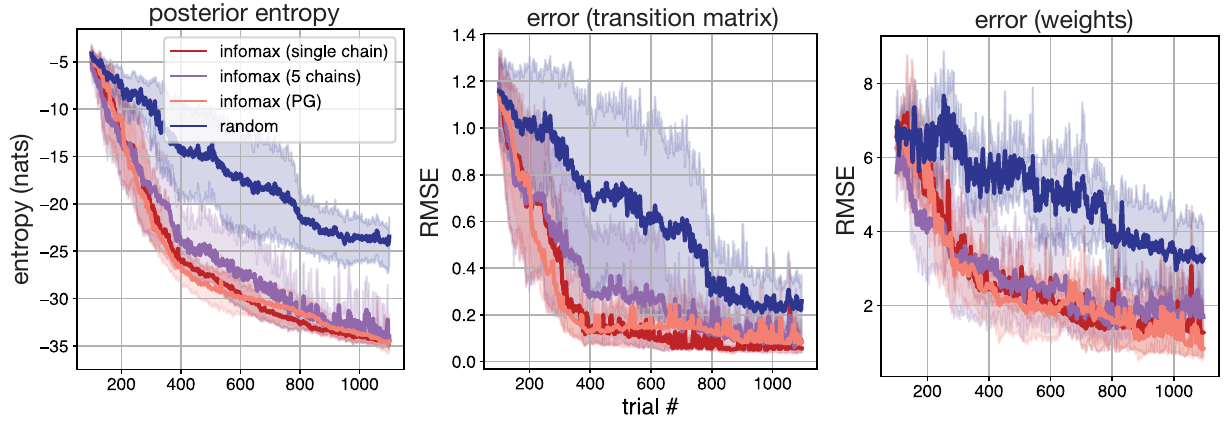}
    \caption{Infomax learning for GLM-HMMs. Left panel shows the posterior entropy of model parameters over the course of 1000 trials when performing infomax learning using our Laplace-based Gibbs sampling approach with a single long chain (red), using parallel chains of our Laplace-based Gibbs sampler (violet), using Polya-Gamma augmented Gibbs sampling (peach), and using random sampling (blue). Middle and right panels show error in recovering the transition matrix and the weights of the GLMs using the same set of methods.}
    \label{fig:iohmm_gibbs_variant}
\end{figure*}

Here, we compare our infomax learning method using variants of Gibbs sampling. In all our experiments in sec.~\ref{sec:iohmmsexps}, we run a single chain to obtain $500$ samples of the model's parameters, discarding the initial 200 burn-in samples. If we instead run 5 parallel chains, each of length $140$ and discard the first 40 samples as burn-in, we would still be able to obtain $500$ samples of the model parameters to perform infomax learning, but this provides a 5X improvement in speed, leading to $\sim 4$ secs per trial for input selection. We verify in Fig.~\ref{fig:iohmm_gibbs_variant} that the perform of infomax while using parallel chains of Gibbs is comparable to that using a single long chain (compare the red and violet traces).

Finally, in all our experiments, we use our Laplace-based Gibbs sampling approach for GLM-HMMs (detailed in sec.~\ref{sec:gibbs_iohmm}). We compared this to Polya-Gamma augmented Gibbs sampling \cite{Polson2013pg,pillow_fully_2012}, an established technique in the literature to sample from logistic models. In this case, weights of the GLM are sampled using Polya-Gamma augmentation, while the strategy for sampling the latents and the state transitions remain the same as in algorithm~\ref{alg:iohmmgibbs}. We show in Fig.~\ref{fig:iohmm_gibbs_variant} that our approach is comparable to Polya-Gamma augmentation in terms of both posterior entropy and error in recovering the model parameters (compare the peach and red curves). This empirically verifies the utility of our Laplace-based Gibbs sampling approach for GLM-HMMs.

\section{Gibbs Sampling for MGLMS} \label{sec:gibbs_moglm}
Gibbs sampling for MGLMs is similar to that for GLM-HMMs except that now the states can be sampled independently of each other. Alg.~\ref{alg:moglmgibbs} provides full details. We set a Dirichlet prior over the initial state distribution, with $\alpha_0 = \mathbf{1} \in \mathbb{R}^K$, and that over the weights to be $\Nrm(\vzero,10)$. Here, we run Gibbs sampling for $700$ iterations and discard the first 200 as burn-in (MGLMs require a longer burn-in period).
\begin{algorithm}[h]
  \caption{MGLMs Gibbs Sampling}
  \label{alg:moglmgibbs}
    \begin{algorithmic}[1]
      \State \textbf{Input}: Observations $y_{1:T}$, Inputs $\vx_{1:T}$, Priors: $\alpha_0$, $\vw_0$, $\sigma_0$
      \State \textbf{Output}: Samples $\{({z}_{1:T}$, ${\vw}_{1:K}$, $\pi)^{(j)} \}$
      \State Initialize $z_{1:T}$, $\vw_{1:K}$, $A$, $\pi$
      \For{$j \gets 1,...M$}
        \For{$k \gets 1,...K$}
            \State $\vw^j_k \gets$ \Call{GLMsampleposterior}{\{$y_t, \vx_t \mid z_t=k \}_{1:T}, \vw_{0},\sigma_0, \vw^{j-1}_k$}
        \EndFor
        \State $\pi^j \gets \sample \text{Dir}(\alpha_0 + \mathbf{n})$ 
             \Comment{where $n_{k} = \sum_t \mathbf{I}(z_t=k)$}
        \State $z_t^j \gets \sample P(z_t \mid y_t, \vx_t) \forall \ t = \{1:T\}$ \Comment{s.t. $P(z_t = k\mid y_t, \vx_t) = \frac{P(y=y_t\mid \vx_t, \textbf{w}_k)
      \pi_k}{ \sum_k P(y=y_t\mid \vx_t, \textbf{w}_k) \pi_k}$} 
      \EndFor
    \end{algorithmic}
\end{algorithm}

% \section{Code}\label{sec:code}
% We have attached a zipped version of our code for active learning with MLRs, GLM-HMMs as well as MGLMs. We built our code for GLM-HMMs on top of the SSM toolbox \cite{linderman_ssm_2020}.

\end{document}